\documentclass{article}

\usepackage{microtype}
\usepackage{graphicx}
\usepackage{booktabs} 
\usepackage{longtable}

\usepackage{hyperref}

\usepackage[accepted]{icml2024}

\usepackage{amsmath}
\usepackage{amssymb}
\usepackage{mathtools}
\usepackage{amsthm}
\usepackage{subcaption}
\usepackage{caption}

\usepackage{hyperref}
\usepackage{url}
\usepackage{booktabs}       
\usepackage{amsfonts}       
\usepackage{nicefrac}       
\usepackage{microtype}      
\usepackage{xcolor}         
\usepackage{graphicx}
\usepackage{wrapfig}
\usepackage{multirow}
\usepackage{epsfig}
\usepackage{marvosym}
\usepackage{color}
\usepackage{threeparttable}
\usepackage{setspace}
\usepackage{verbatim}
\usepackage{bbm}
\usepackage{bm}
\usepackage{comment}
\usepackage{pifont}
\usepackage{nicematrix}
\newcommand{\cmark}{\ding{51}}%
\newcommand{\xmark}{\ding{55}}%
\newcommand{\bcmark}{\textbf{\cmark}}
\newcommand{\bxmark}{\textbf{\xmark}}

\usepackage{xspace}
\newcommand{\abb}{\texttt{TSLANet}\xspace}

\newcommand{\boldres}[1]{{\textbf{\textcolor{blue}{#1}}}}
\newcommand{\secondres}[1]{{\underline{\textcolor{purple}{#1}}}}
\usepackage{listings}

\icmltitlerunning{TSLANet: Rethinking Transformers for Time Series Representation Learning}

\begin{document}

\twocolumn[
\icmltitle{TSLANet: Rethinking Transformers for Time Series Representation Learning}

\begin{icmlauthorlist}
\icmlauthor{Emadeldeen Eldele}{cfar}
\icmlauthor{Mohamed Ragab}{cfar,i2r}
\icmlauthor{Zhenghua Chen}{cfar,i2r}
\icmlauthor{Min Wu}{i2r}
\icmlauthor{Xiaoli Li}{cfar,i2r}
\end{icmlauthorlist}

\icmlaffiliation{cfar}{Centre for Frontier AI Research, Agency for Science, Technology and Research, Singapore}
\icmlaffiliation{i2r}{I2R, Agency for Science, Technology and Research, Singapore}

\icmlcorrespondingauthor{Emadeldeen Eldele}{emad0002@ntu.edu.sg}

\icmlkeywords{Time Series, Classification, Forecasting, Fourier Transform, Convolutional Neural Networks}

\vskip 0.3in
]

\printAffiliationsAndNotice{}  

\begin{abstract}
Time series data, characterized by its intrinsic long and short-range dependencies, poses a unique challenge across analytical applications. While Transformer-based models excel at capturing long-range dependencies, they face limitations in noise sensitivity, computational efficiency, and overfitting with smaller datasets. In response, we introduce a novel \textbf{T}ime \textbf{S}eries \textbf{L}ightweight \textbf{A}daptive \textbf{Net}work (\abb), as a universal convolutional model for diverse time series tasks. Specifically, we propose an Adaptive Spectral Block, harnessing Fourier analysis to enhance feature representation and to capture both long-term and short-term interactions while mitigating noise via adaptive thresholding. Additionally, we introduce an Interactive Convolution Block and leverage self-supervised learning to refine the capacity of \abb for decoding complex temporal patterns and improve its robustness on different datasets. Our comprehensive experiments demonstrate that \abb outperforms state-of-the-art models in various tasks spanning classification, forecasting, and anomaly detection, showcasing its resilience and adaptability across a spectrum of noise levels and data sizes. The code is available at \url{https://github.com/emadeldeen24/TSLANet}.
\end{abstract}

\section{Introduction}
\label{sec:intro}
\begin{figure}[t]
    \centering
    \includegraphics[width=0.9\columnwidth]{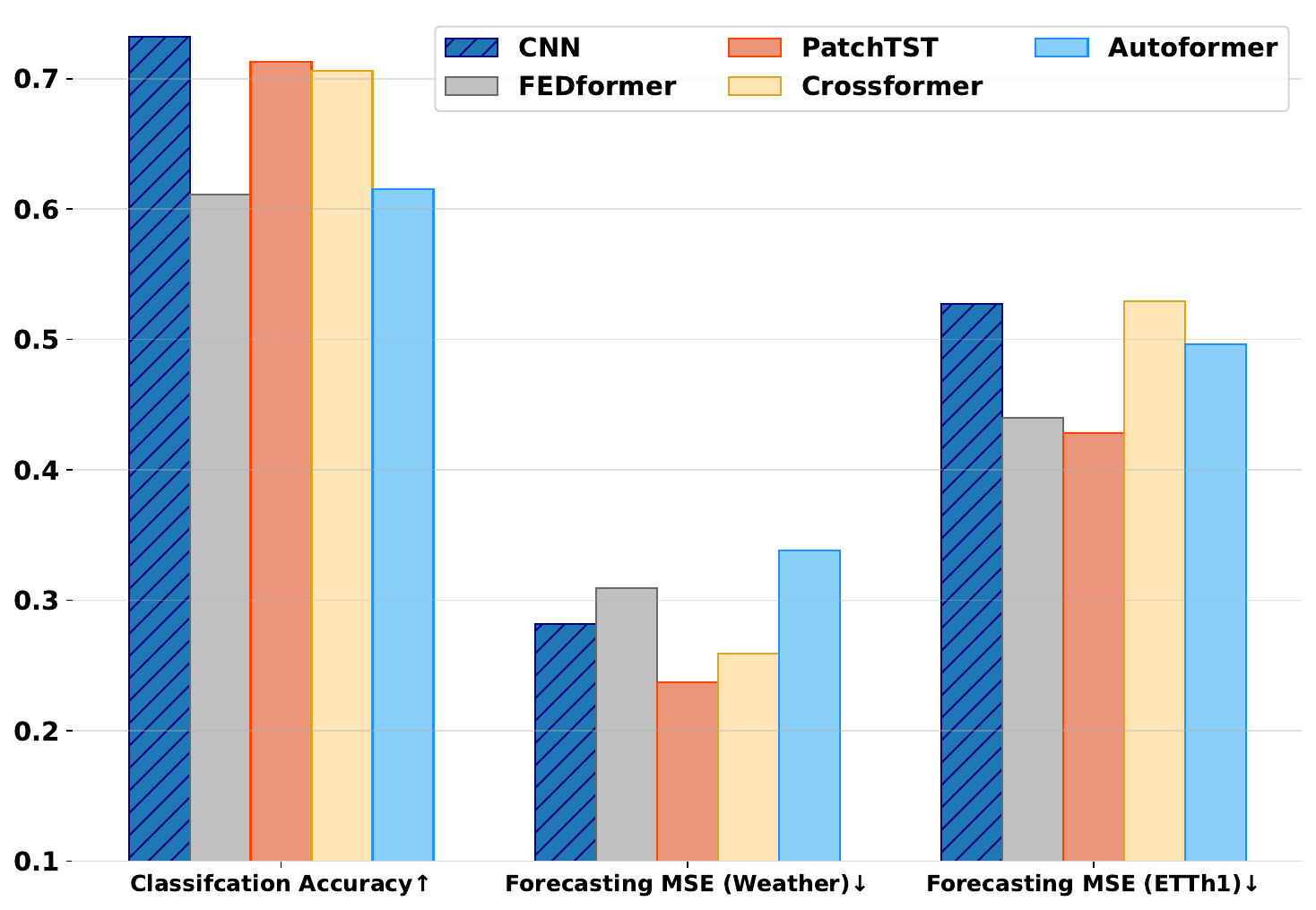}
    \caption{A comparison between CNN and Transformer-based architectures for classification and forecasting tasks. Classification results are the average over 10 UEA datasets \cite{Timesnet}, while forecasting results are the average MSE results on lengths \{96, 192, 336, 720\}.}
    \label{fig:cnn_vs_transformers}
\end{figure}

Time series data, known for its sequential nature and temporal dependencies, is ubiquitous across numerous domains, including finance, healthcare, and environmental monitoring. 
Recently, the Transformer model \cite{Transformer}, originally renowned for its breakthroughs in natural language processing, has been adapted as a potent tool for analyzing time series data. 
This was motivated by its ability to capture long-range dependencies and interactions within time series data, showing proficiency in forecasting tasks \cite{Autoformer,fedformer,itransformer}.
Despite the initial success of Transformers in time series forecasting, they encounter hurdles when deployed across diverse time series tasks, particularly those with smaller datasets. This can be attributed to its large parameter size, which may lead to overfitting and computational inefficiency problems \cite{transformer_ts_survey}. In addition, their attention mechanism often struggles with the inherent noise and redundancy in time series data \cite{uniformer}. Moreover, recent works have questioned their adaptability, as highlighted by \citep{DLinear,RLinear}. They observed that the self-attention within Transformers is inherently permutation-invariant, which compromises the preservation of temporal information. Their experiments showed that a single linear layer surprisingly outperforms the complex Transformer architectures for time series forecasting. However, while such linear models can perform well for small, clean data, they may not be able to handle complex, noisy time series.

In this work, we pivot from the prevalent focus on  Multi-Layer Perceptrons (MLPs) and Transformers to tackle the potential of convolutional operations for time series analysis. Convolutional Neural Networks (CNNs) have traditionally excelled in capturing short-term patterns within time series due to their local receptive fields, which serve as a strength in classification tasks. Indeed, as illustrated in Figure~\ref{fig:cnn_vs_transformers}, a straightforward 3-layer CNN network demonstrates superior performance in classification compared to state-of-the-art Transformer-based architectures.
Yet, our experiment showed that the efficacy of CNNs in forecasting varies with the data frequency. For instance, the CNN shows competitive performance to these Transformer-based models on the Weather dataset featuring a short 10-minute frequency but struggles with the longer hourly ETTh1 dataset, indicating a difficulty with less frequent temporal changes. This discrepancy highlights a critical question: \textit{How can we enhance CNNs to extend their robust performance across a wider spectrum of time series tasks?} It becomes obvious that expanding the capabilities of CNNs can be achieved by learning both short-term and long-term dependencies within time series data.

To this end, we introduce \textbf{T}ime \textbf{S}eries \textbf{L}ightweight \textbf{A}daptive \textbf{Net}work (\abb), a universal architecture for various time series tasks. \abb inherits the multi-block design of the Transformer to allow scalability. However, we replace the computationally expensive self-attention with a lightweight Adaptive Spectral Block (ASB) featuring two key objectives. Firstly, ASB aims to encompass the entire frequency spectrum, thereby adeptly capturing both long-term and short-term interactions within the data. This is achieved via Fourier-based multiplications by global and local filters, akin to circular convolutions. Secondly, ASB selectively attenuates high frequencies via an adaptive thresholding approach, a strategy aimed at minimizing noise and enhancing the clarity of the signal. 
In addition, we further advance our model by replacing the standard feed-forward network with an Interactive Convolutional Block, where CNNs with different kernel sizes control each other to enrich the ability of the model to capture and interpret complex temporal patterns. Finally, we employ a per-dataset self-supervised pretraining to enhance the model capabilities, especially on large datasets.

The proposed model is lightweight and enjoys the $\mathcal{O}(N\log N)$ complexity of the Fast Fourier Transform (FFT) operations, demonstrating superior efficiency and speed compared to self-attention (see Section \ref{sec:complexity_analysis}). A summary comparison against CNN-based and Transformer-based models is also provided in Table~\ref{tbl:high_level_comparison}. The contributions of this paper can be summarized as follows:
\begin{itemize}
    \item We propose a universal lightweight model (\abb), designed to adapt seamlessly to a myriad of time series tasks. Through computationally efficient convolution operations, \abb learns both long- and short-term relationships within the data.
    
    \item We propose an Adaptive Spectral Block, which leverages the power of Fourier transform alongside global and local filters to cover the whole frequency spectrum, while adaptively removing high frequencies that tend to introduce noises. In addition, we propose an Interactive Convolution Block to learn intricate spatial and temporal features within data.

    \item \abb demonstrates superior performance against different state-of-the-art methods across various time series tasks.
    
\end{itemize}

\begin{table}[!t]
\centering
\caption{Comparison to different methods. `Local Dependencies' means the efficiency in capturing local features.}
\label{tbl:high_level_comparison}
\resizebox{0.95\columnwidth}{!}{
\begin{NiceTabular}{l|cccc}
\toprule
\textbf{Method} & \textbf{Feature Extraction} & \begin{tabular}[c]{@{}c@{}}\textbf{Long-range}\\ \textbf{Dependencies}\end{tabular} & \begin{tabular}[c]{@{}c@{}}\textbf{Local}\\ \textbf{Dependencies}\end{tabular} & \begin{tabular}[c]{@{}c@{}}\textbf{Parameter}\\ \textbf{Efficiency}\end{tabular}\\ \midrule
CNN  & Localized Convolution  & \bxmark & \bcmark & \bcmark \\
Transformer   & Self-Attention   & \bcmark & \bxmark   & \bxmark \\
\abb         & Adaptive Spectral Convolution & \bcmark    & \bcmark & \bcmark  \\ \bottomrule
\end{NiceTabular}
}
\vskip 0.05in
\end{table}

\begin{figure*}
    \centering
    \includegraphics[width=0.75\textwidth]{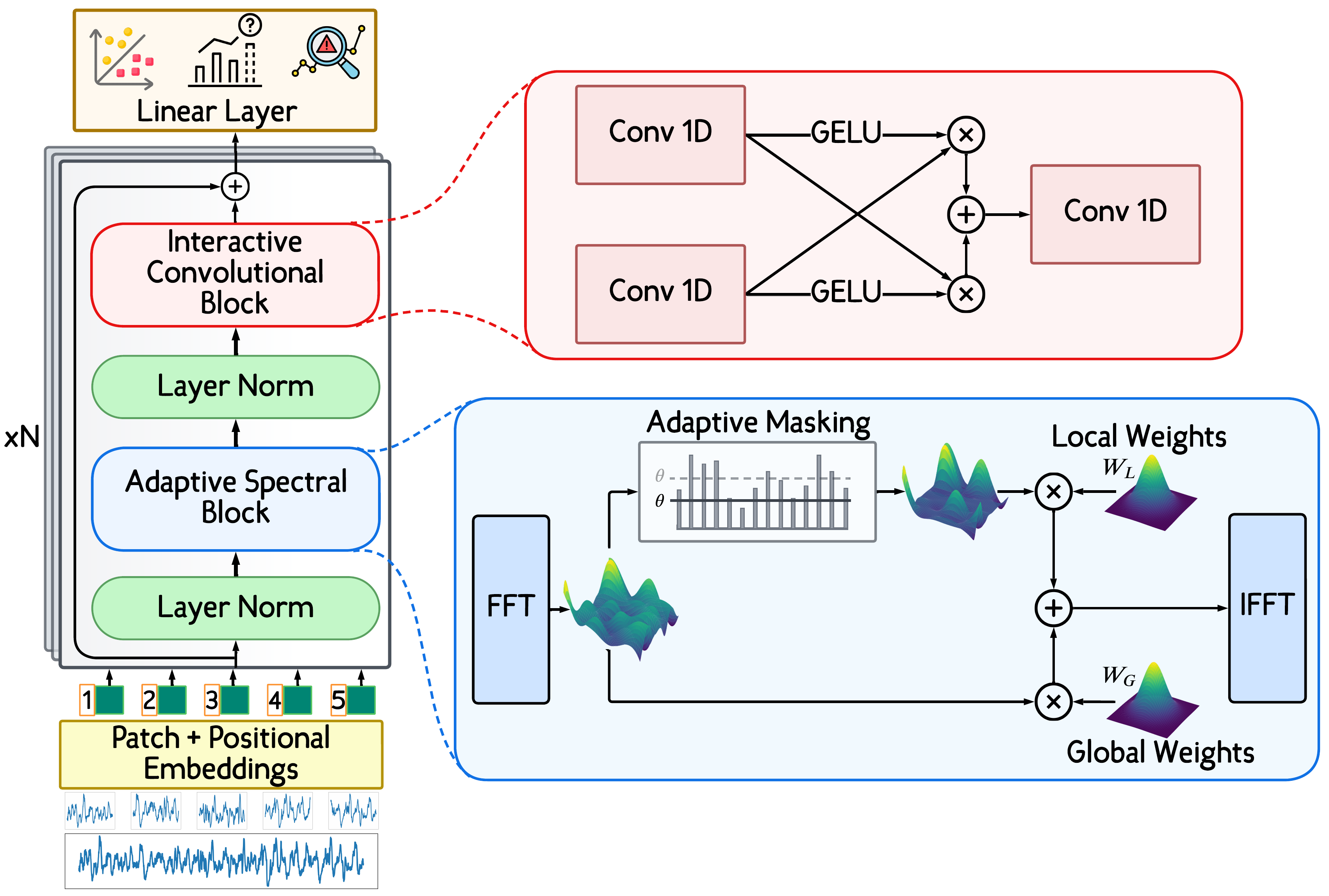}
    \caption{The structure of our proposed \abb. The input time series is split into patches, and positional embeddings are added. Next, the output embeddings pass through \abb layers, where each layer consists of two main components. The first is the Adaptive Spectral Block, which leverages frequency domain representations for robust feature extraction and employs adaptive thresholding to mitigate noise. The second is the Interactive Convolution Block, which captures complex temporal patterns through convolutional operations.}
    \label{fig:main}
\end{figure*}

\section{Related Works}
\paragraph{Transformer-based Networks.}
Since the advance of the Transformer~\cite{Transformer} for natural language processing, numerous works have adopted it for time series analysis. For example, \cite{Autoformer,fedformer,Informer,reformer,Crossformer} have showcased the Transformer capability to model interactions within time series data, utilizing that for the forecasting task. In addition, Transformers with special design showed good performance in anomaly detection task \cite{xu2021anomaly}.

Yet, the efficacy of Transformers for time series has been contested. For instance, \citet{DLinear} argue that the permutation-invariance property in Transformers may lead to the loss of temporal information in time series. Following that, other MLP-based architectures showed efficacy in the time series forecasting task \cite{RLinear,TSMixer}. Furthermore, Transformers demand extensive computational resources in general, and they are prone to overfitting when trained on smaller datasets \cite{transformer_ts_survey}.

\paragraph{Convolution-based Networks.}
CNNs have showcased their efficacy in time series analysis, particularly shining in classification tasks due to their adeptness at learning local patterns \cite{ROCKET}. CNNs also serve as the backbone for several time series representation learning methods, including TS-TCC \cite{tstcc}, TS2VEC \cite{ts2vec}, and MHCCL \cite{mhccl}.

Despite their promise, CNNs often face challenges in forecasting and anomaly detection, primarily due to their limited ability to capture long-range dependencies. Therefore, recent works attempt to enhance CNN capabilities in different ways. For instance, T-WaveNet \cite{tWaveNet} leverages frequency spectrum energy analysis for effective signal decomposition, SCINet \cite{SCINet} adopts a recursive downsample-convolve-interact strategy to model complex temporal dynamics, and WFTNet \cite{WFTNet} employs a combination of Fourier and wavelet transforms for a thorough temporal-frequency analysis. Additionally, TCE \cite{TCE} targets the improvement of 1D-CNNs by addressing the disturbing convolution for better low-frequency component focus, and BTSF \cite{pmlr-v162-yang22e} introduces a bilinear temporal-spectral fusion technique for unsupervised learning, emphasizing the importance of maintaining the global context of time series data.

A noteworthy attempt to leverage CNNs for multiple time series tasks is the TimesNet model \cite{Timesnet}, which capitalizes on multi-periodicity to merge intraperiod and interperiod variations within a 2D space, enhancing the representation of temporal patterns. However, TimesNet may not fully address the challenges presented by non-stationary datasets lacking clear periodicity. Some recent works have explored combining CNNs with Transformers to harness both their strengths \cite{uniformer,cvt,ConViT}, though such hybrid approaches remain underexplored in time series analysis compared to their applications in computer vision.

Our work takes a distinct path by proposing a universal convolutional-based architecture, adept at handling various time series tasks through adaptive spectral feature extraction. This approach not only utilizes the strong local feature learning capabilities of CNNs but also efficiently captures global temporal patterns, offering a balanced solution for both local and long-range dependencies in time series data.

\section{Method}
\label{sec:method}
\subsection{Preliminaries: Discrete Fourier Transform}
We first explore the Discrete Fourier Transform (DFT) as it is a cornerstone in our framework. Consider a series of $N$ complex numbers $x[n]$, where $0 \leq n \leq N-1$. The 1D DFT transforms this series into a frequency domain representation:
\begin{equation}
X[k] = \sum_{n=0}^{N-1} x[n] e^{-j(2\pi / N) k n} := \sum_{n=0}^{N-1} x[n] W_N^{kn},
\label{equ:dft}
\end{equation}
where $j$ denotes the imaginary unit, with $W_N = e^{-j(2\pi/N)}$. This formulation is derived from the continuous Fourier transform by discretizing in both time and frequency domains. The spectrum of the sequence $x[n]$ at frequency $\omega_k = 2\pi k/N$ is represented by $X[k]$, which is periodic with an interval of length $N$, thus only the first $N$ points are considered.

Due to the bijective nature of DFT, the original sequence $x[n]$ is retrievable via the Inverse DFT (IDFT):
\begin{equation}
x[n] = \frac{1}{N} \sum_{k=0}^{N - 1}X[k]e^{j(2\pi/N)kn}.
\label{equ:idft}
\end{equation}
For real-valued $x[n]$, DFT exhibits conjugate symmetry, i.e., $X[N - k] = X^*[k]$. This symmetry is pivotal, as performing IDFT on a conjugate symmetric $X[k]$ results in a real discrete signal. Half of the DFT spectrum, specifically ${X[k]: 0\le k\le \lceil N / 2\rceil}$, sufficiently describes the frequency characteristics of $x[n]$.

The choice of DFT in \abb is motivated by two factors: its discrete nature aligns well with digital processing and the existence of efficient computation methods. The Fast Fourier Transform (FFT), leveraging the symmetry and periodicity of $W_N^{kn}$, optimizes DFT computation from $\mathcal{O}(N^2)$ to $\mathcal{O}(N\log N)$. The IDFT, paralleling DFT's form, benefits similarly from the Inverse FFT (IFFT).

\subsection{Overall Architecture} 
Our model integrates two novel components, i.e., the Adaptive Spectral Block (ASB) and the Interactive Convolution Block (ICB), as depicted in Figure~\ref{fig:main}. These two components form a single layer that could be extended to multiple layers. The ASB employs Fourier analysis to transform time series data into the frequency domain, in which we apply adaptive thresholding to attenuate high-frequency noise and highlight relevant spectral features. After processing, the IFFT reconstructs the time-domain features, now with reduced noise and enhanced representations. The ICB is a streamlined convolutional block that interactively refines features using different kernel sizes, improving adaptability to temporal dynamics in time series. Together, these components form a cohesive structure that balances local and global temporal feature extraction for time series analysis.

\subsection{Embedding Layer}
Given an input time series \( \bm{S} \), with each signal \( S \in \mathbb{R}^{C \times L} \) having \( C \) channels and a sequence length \( L \). First, the signal \( S \) is divided into a set of $M$ patches \( \{P_1, P_2, ..., P_M\} \), where each patch \( P_i \) captures a segment of \( S \). The dimension of each patch is determined by the predefined patch size \( p \), such that each patch \( P_i \in \mathbb{R}^{C \times p}\).

Each patch is then mapped into another dimension \( p' \), i.e.,  \( P_i \rightarrow P_i' \in \mathbb{R}^{C \times p'}\). Next, the positional embeddings are added to each patch to retain the temporal ordering disrupted during the segmentation process. The positional embedding for the \(i\)-th patch is denoted as \( E_i \), a vector that aligns dimensionally with the patch. The augmented patch results from adding both inputs, i.e., $S_{PE_{i}} = P_i' + E_i$, and $S_{PE} = \{ S_{PE_{1}}, S_{PE_{2}}, \dots S_{PE_{M}}\}$. Notably, the positional embeddings are learnable parameters, allowing the model to capture the temporal relationships within the time series data effectively.

\subsection{Adaptive Spectral Block}
We propose the Adaptive Spectral Block (ASB) that employs the Fourier-domain processing, as inspired by \cite{GFNets}. This block aims to learn spatial information with the global circular convolution operations. 
Moreover, it provides adaptive local filters to isolate noisy high-frequency components for any time series data.

\paragraph{Fast Fourier Transformations.} 
Given a discrete time series \( x[n] \), we obtain its frequency domain representation  \( X[k] \), by performing FFT along the spatial dimensions as in Equation~\ref{equ:dft}. Similarly, given \( \bm{S_{PE}} \), its representation is calculated as:
\begin{equation}
\bm{F} = \mathcal{F}[\bm{S_{PE}}] \in \mathbb{C}^{C \times L'},
\end{equation}
where \( \mathcal{F}[\cdot] \) denotes the 1D FFT operation, and \( L' \) is the transformed sequence length in the frequency domain, which may differ from \( L \) depending on the FFT implementation and the nature of the time series data. Each channel of the time series is independently transformed, resulting in a comprehensive frequency domain representation \( \bm{F} \) that encapsulates the spectral characteristics of the original time series across all channels.

\paragraph{Adaptive Removal of High-Frequency Noise.}
High-frequency components often represent rapid fluctuations that deviate from the underlying trend or signal of interest, making them appear more random and difficult to interpret \cite{app9071345}. 
Therefore, we propose an adaptive local filter that allows the model to \textit{dynamically} adjust the level of filtering according to the dataset characteristics and remove these high-frequency noisy components. This is crucial when dealing with non-stationary data, where the frequency spectrum may change over time. The proposed filter adaptively sets the appropriate frequency threshold for each specific time series data.

Given the frequency domain representation \( \bm{F} \) obtained from the FFT operation, we first calculate the power spectrum of \( \bm{F} \), which helps in identifying dominant frequency components. The power spectrum \( \bm{P} \) is computed as the square of the magnitude of the frequency components:  \( \bm{P} = |\bm{F}|^2 \), which gives us a measure of the strength of different frequencies in the time series data.

The key to effective noise reduction lies in adaptively filtering high-frequency components from the power spectrum \( \bm{P} \). We achieve this with a trainable threshold \( \theta \), which adjusts based on the spectral characteristics of the data. This threshold \( \theta \) is set as a learnable parameter optimized during training through backpropagation, specifically \( \frac{\partial \mathcal{L}}{\partial \theta} \), enabling \( \theta \) to discern between essential signal frequencies and noise. We formulate this adaptive thresholding as follows:
   \begin{equation}
   \bm{F}_{\text{filtered}} = \bm{F} \odot (\bm{P} > \theta),
   \end{equation}
where \( \odot \) represents element-wise multiplication, and \( (\bm{P} > \theta) \) is a binary mask where frequencies with power above the threshold \( \theta \) are retained, and others are filtered out.

The adaptability of the threshold \( \theta \) ensures that the ASB can efficiently remove high frequencies while preserving crucial information. By adaptively selecting the frequency threshold, the ASB tailors its filtering process to each specific time series dataset, enhancing the overall effectiveness of the model in handling a wide range of data scenarios.

\paragraph{Learnable Filters.}
After adaptively filtering the frequency domain data, the model employs two sets of learnable filters; a global filter to learn from the original frequency domain data \( \bm{F} \) and a local filter to learn from the adaptively filtered data \( \bm{F}_{\text{filtered}} \). Let \( \bm{W}_{\text{G}} \) and \( \bm{W}_{\text{L}} \) be the learnable global and local filters, respectively. The application of these filters is represented as:
   \begin{align}
   \bm{F}_{\text{G}} &= \bm{W}_{\text{G}} \odot \bm{F},
   \label{eqn:global_filter_mult} \\
   \bm{F}_{\text{L}} &= \bm{W}_{\text{L}} \odot \bm{F}_{\text{filtered}}.
   \label{eqn:local_filter_mult}
   \end{align}
Next, we integrate these filtered features to capture a comprehensive spectral detail, i.e.,  $\bm{F}_{\text{integrated}} = \bm{F}_{\text{G}} + \bm{F}_{\text{L}}$.









\begin{table*}[!h]
\caption{Classification results in different datasets. Results are averaged across each subset of datasets. Results are in terms of accuracy (as \%). \boldres{Blue}: best results, \secondres{Purple}: second best. Full results are listed in Tables \ref{tbl:clf_ucr_full}, \ref{tbl:clf_uea_full}, and \ref{tbl:clf_others_full} in the Appendix.}
\label{tbl:classification_short}
\scalebox{0.8}{
\setlength\tabcolsep{3pt}
\begin{threeparttable}[b]
\begin{NiceTabular}{c|ccccccccc}
\toprule

\multirow{2}{*}{Methods} 
& \abb & GPT4TS & TimesNet & ROCKET & Crossformer & PatchTST & MLP & TS-TCC & TS2VEC\\
&\textbf{\textit{(Ours)}} & \citeyearpar{onefitsall} & \citeyearpar{Timesnet} & \citeyearpar{ROCKET} & \citeyearpar{Crossformer} & \citeyearpar{PatchTST} & \citeyearpar{DLinear} & \citeyearpar{tstcc} & \citeyearpar{ts2vec} \\

\midrule
UCR repository (85 datasets) & \boldres{83.18} & 61.58 & 65.27 & \secondres{81.42} & 73.47 & 71.84 & 69.68 & 75.07 & \secondres{81.42} \\
UEA repository (26 datasets) & \boldres{72.73} & 58.51 & 66.55 & 68.79 & 66.84 & 69.13 & 65.81 & \secondres{69.38} & 59.62 \\
Biomedical signals (2 datasets) & \secondres{90.24} & 87.04 & 87.10 & 87.20 & 70.82 & 83.87 & 70.63 & \boldres{92.25} & 86.31 \\
Human activity recognition (3 datasets) & \boldres{97.46} & 92.71 & 91.51 & 96.44 & 77.55 & 94.87 & 56.69 & \secondres{97.16} & 95.70 \\
\midrule
Average & \boldres{85.90} & 74.96 & 77.61 & 83.46 & 72.17 & 79.93 & 65.70 & \secondres{83.55} & 80.76 \\
\bottomrule
\end{NiceTabular}
\end{threeparttable}
}
\end{table*}

Notably, the multiplication operations in Equations \ref{eqn:global_filter_mult} and \ref{eqn:local_filter_mult} are equivalent to the circular convolution process (see Appendix~\ref{sec:appen1:cir_conv}). Circular convolution, with its larger receptive field over the entire sequence, is particularly adept at capturing periodic patterns in time series data.

\paragraph{Inverse Fourier Transform.} 
To convert the integrated frequency domain data back to the time domain, we apply the Inverse Fast Fourier Transform (IFFT). The resulting time-domain signal \( \bm{S}' \) is given by:
   \begin{equation}
   \bm{S}' = \mathcal{F}^{-1}[\bm{F}_{\text{integrated}}] ~~~~~\in \mathbb{R}^{C \times p'}.
   \end{equation}
The IFFT ensures that the enhanced features align with the original data structure of the input time series. The full operation of the ASB is described in Algorithm~\ref{alg:adaptive_global_filter} in the Appendix.

\subsection{Interactive Convolution Block}
After enhancing feature representation by the ASB, we propose the Interactive Convolution Block (ICB), which utilizes a dual-layer convolutional structure, as shown in Figure~\ref{fig:main}. The design of the ICB includes parallel convolutions with different kernel sizes to capture local features and longer-range dependencies.
Specifically, the first convolutional layer is designed to capture fine-grained, localized patterns in the data with a smaller kernel. In contrast, the second layer aims to identify broader, longer-range dependencies with a larger kernel. 
We design the ICB such that the output of each layer modulates the feature extraction of the other. The element-wise multiplication encourages interactions between features extracted at different scales, potentially leading to better modeling of complex relationships.

Given the output of the IFFT operation \( \bm{S}' \), it serves as the input to the ICB. The process within the ICB is as follows:
   \begin{equation}
        \bm{A}_1 = \phi(\text{Conv1}(\bm{S}')) \odot \text{Conv2}(\bm{S}'),
   \end{equation}
   \begin{equation}
        \bm{A}_2 = \phi(\text{Conv2}(\bm{S}')) \odot \text{Conv1}(\bm{S}'),
   \end{equation}
where \(\text{Conv1}(\cdot)\) and \(\text{Conv2}(\cdot)\) are two 1D-convolutional layers and \( \phi \) is the GELU activation function.

The activated features are then added and passed through a final convolutional layer \(\text{Conv3}(\cdot)\):
\begin{equation}
   \bm{O}_{\text{ICB}} = \text{Conv3}(\bm{A}_1 + \bm{A}_2). 
\end{equation}
The output \( \bm{O}_{\text{ICB}} \) represents the enhanced features ready for the final layer in the network, represented by a customizable linear layer according to the task.

\subsection{Self-Supervised Pretraining}
Expanding the capabilities of \abb, we incorporate a phase of self-supervised pretraining, which has garnered significant attention for its efficacy in learning high-level representations from unlabeled data \cite{PatchTST}. Drawing inspiration from methodologies applied in natural language processing and computer vision, we adopt a masked autoencoder paradigm for time series data \cite{mae}.

Our implementation involves selective masking of input sequence patches, followed by training \abb to reconstruct these masked segments accurately. The masked data then serves as the training input, compelling the model to learn and infer the underlying patterns and dependencies in the data. Unlike methods that apply masking at individual time steps, our approach focuses on larger patches. This design choice avoids simplistic interpolation from adjacent time points and encourages the model to understand the entire sequence deeply. The reconstruction of these patches is achieved by optimizing the mean squared error (MSE) loss function.

\section{Experiments}

\begin{table*}[htbp]
  \caption{Multivariate forecasting results with prediction lengths $\in\{96, 192, 336, 720\}$. Results are averaged from all prediction lengths. \emph{Avg means further averaged by subsets}. \boldres{Blue}: best results, \secondres{Purple}: second best. Full results are listed in Table \ref{tbl:forecasting_full} in the Appendix.}
  \label{tbl:forecasting_short}
  \renewcommand{\arraystretch}{0.8} 
  \centering
  \resizebox{0.95\textwidth}{!}{
  \begin{threeparttable}
  \begin{small}
  \renewcommand{\multirowsetup}{\centering}
  \setlength{\tabcolsep}{1.45pt}
  
  \begin{NiceTabular}{c|cc|cc|cc|cc|cc|cc|cc|cc|cc|cc|cc|cc}
    \toprule
        {\multirow{2}{*}{Models}} & 
    \multicolumn{2}{c}{\rotatebox{0}{\scalebox{0.8}{\abb}}} &
    \multicolumn{2}{c}{\rotatebox{0}{\scalebox{0.8}{Time-LLM}}} &
    \multicolumn{2}{c}{\rotatebox{0}{\scalebox{0.8}{iTransformer}}} &
    
    \multicolumn{2}{c}{\rotatebox{0}{\scalebox{0.8}{PatchTST}}} &
    \multicolumn{2}{c}{\rotatebox{0}{\scalebox{0.8}{Crossformer}}} &
    
    \multicolumn{2}{c}{\rotatebox{0}{\scalebox{0.8}{FEDformer}}} &
    
    \multicolumn{2}{c}{\rotatebox{0}{\scalebox{0.8}{Autoformer}}} &
    \multicolumn{2}{c}{\rotatebox{0}{\scalebox{0.8}{RLinear}}} &
    \multicolumn{2}{c}{\rotatebox{0}{\scalebox{0.8}{Dlinear}}} &
    \multicolumn{2}{c}{\rotatebox{0}{\scalebox{0.8}{{TimesNet}}}} &
    \multicolumn{2}{c}{\rotatebox{0}{\scalebox{0.8}{GPT4TS}}} &
    \multicolumn{2}{c}{\rotatebox{0}{\scalebox{0.8}{SCINet}}} \\
     &
    \multicolumn{2}{c}{\scalebox{0.75}{\textit{\textbf{(Ours)}}}} &
    \multicolumn{2}{c}{\scalebox{0.8}{\citeyearpar{timellm}}} &
    \multicolumn{2}{c}{\scalebox{0.8}{\citeyearpar{itransformer}}} &
     
    \multicolumn{2}{c}{\scalebox{0.8}{\citeyearpar{PatchTST}}} & 
    \multicolumn{2}{c}{\scalebox{0.8}{\citeyearpar{Crossformer}}} & 
    \multicolumn{2}{c}{\scalebox{0.8}{\citeyearpar{fedformer}}} &
    \multicolumn{2}{c}{\scalebox{0.8}{\citeyearpar{Autoformer}}} &
    \multicolumn{2}{c}{\scalebox{0.8}{\citeyearpar{RLinear}}} &
    \multicolumn{2}{c}{\scalebox{0.8}{\citeyearpar{DLinear}}} &
    \multicolumn{2}{c}{\scalebox{0.8}{\citeyearpar{Timesnet}}} & 
    \multicolumn{2}{c}{\scalebox{0.8}{\citeyearpar{onefitsall}}} &
    \multicolumn{2}{c}{\scalebox{0.8}{\citeyearpar{SCINet}}} \\
    \cmidrule(lr){2-3} \cmidrule(lr){4-5}\cmidrule(lr){6-7} \cmidrule(lr){8-9}\cmidrule(lr){10-11}\cmidrule(lr){12-13} \cmidrule(lr){14-15} \cmidrule(lr){16-17} \cmidrule(lr){18-19} \cmidrule(lr){20-21} \cmidrule(lr){22-23} \cmidrule(lr){24-25}
    {Metric} & \scalebox{0.8}{MSE} & \scalebox{0.8}{MAE} & \scalebox{0.8}{MSE} & \scalebox{0.8}{MAE}  & \scalebox{0.8}{MSE} & \scalebox{0.8}{MAE}  & \scalebox{0.8}{MSE} & \scalebox{0.8}{MAE}  & \scalebox{0.8}{MSE} & \scalebox{0.8}{MAE}  & \scalebox{0.8}{MSE} & \scalebox{0.8}{MAE}  & \scalebox{0.8}{MSE} & \scalebox{0.8}{MAE} & \scalebox{0.8}{MSE} & \scalebox{0.8}{MAE} & \scalebox{0.8}{MSE} & \scalebox{0.8}{MAE} & \scalebox{0.8}{MSE} & \scalebox{0.8}{MAE} & \scalebox{0.8}{MSE} & \scalebox{0.8}{MAE} & \scalebox{0.8}{MSE} & \scalebox{0.8}{MAE} \\
    \toprule
    
    \scalebox{0.95}{ECL} 
    & \secondres{\scalebox{0.8}{0.165}} & \secondres{\scalebox{0.8}{0.257}} 
    & \boldres{\scalebox{0.8}{0.158}} & \boldres{\scalebox{0.8}{0.252}} 
    & {\scalebox{0.8}{0.178}} & {\scalebox{0.8}{0.270}} 
    
    & {\scalebox{0.8}{0.167}} & {\scalebox{0.8}{0.259}}
    & \scalebox{0.8}{0.244} & \scalebox{0.8}{0.334}    
    &\scalebox{0.8}{0.214} &\scalebox{0.8}{0.327} 
    
    &\scalebox{0.8}{0.227} &\scalebox{0.8}{0.338} 
    &\scalebox{0.8}{0.219} &\scalebox{0.8}{0.298} 
    &{\scalebox{0.8}{0.166}} &{\scalebox{0.8}{0.263}}  
    &\scalebox{0.8}{0.192} &\scalebox{0.8}{0.295} 
    &\scalebox{0.8}{0.167} &\scalebox{0.8}{0.263}
    & \scalebox{0.8}{0.268} & \scalebox{0.8}{0.365} \\
    \midrule

    \scalebox{0.95}{ETT (Avg)} 
    & \secondres{\scalebox{0.8}{0.337}} & \secondres{\scalebox{0.8}{0.377}} 
    & \boldres{\scalebox{0.8}{{0.330}}} & \boldres{\scalebox{0.8}{{0.372}}}
    & {\scalebox{0.8}{0.383}} & {\scalebox{0.8}{0.399}} 
     
    & {\scalebox{0.8}{0.347}} & {\scalebox{0.8}{0.378}} 
    & \scalebox{0.8}{{0.685}} & \scalebox{0.8}{{0.578}} 
    & \scalebox{0.8}{{0.408}} & \scalebox{0.8}{{0.428}} 
    
    & \scalebox{0.8}{{0.465}} & \scalebox{0.8}{{0.459}}
    & {\scalebox{0.8}{{0.380}}} & {\scalebox{0.8}{0.392}}
    & \scalebox{0.8}{{0.369}} & \scalebox{0.8}{{0.398}} 
    & \scalebox{0.8}{{0.391}} & \scalebox{0.8}{{0.404}} 
    & \scalebox{0.8}{{0.350}} & \scalebox{0.8}{{0.382}} 
    & \scalebox{0.8}{{0.689}} & \scalebox{0.8}{{0.597}} \\
    \midrule 

    \scalebox{0.95}{Exchange} 
    & {\scalebox{0.8}{0.369}} & {\scalebox{0.8}{0.404}}
    & {\scalebox{0.8}{-}} & {\scalebox{0.8}{-}} 
    & \secondres{\scalebox{0.8}{0.360}} & \secondres{\scalebox{0.8}{0.403}} 
    & {\scalebox{0.8}{0.367}} & {\scalebox{0.8}{0.404}} 
    & \scalebox{0.8}{0.940} & \scalebox{0.8}{0.707} 
    &{\scalebox{0.8}{0.519}} &\scalebox{0.8}{0.429} 
    &\scalebox{0.8}{0.613} &\scalebox{0.8}{0.539} 
    &\scalebox{0.8}{0.378} &\scalebox{0.8}{0.417} 
    &\boldres{\scalebox{0.8}{0.297}} &\boldres{\scalebox{0.8}{0.378}} 
    &{\scalebox{0.8}{0.416}} &{\scalebox{0.8}{0.443}} 
    & {\scalebox{0.8}{0.370}} &\scalebox{0.8}{0.406}  
    & \scalebox{0.8}{0.750} & \scalebox{0.8}{0.626}  \\

    \midrule 
    \scalebox{0.95}{Traffic} 
    & \secondres{\scalebox{0.8}{0.396}} & \secondres{\scalebox{0.8}{0.271}}
    & \boldres{\scalebox{0.8}{0.388}} & \boldres{\scalebox{0.8}{0.264}}
    & \scalebox{0.8}{0.428} & {\scalebox{0.8}{0.282}}
    
    & {\scalebox{0.8}{0.420}} & {\scalebox{0.8}{0.277}}
    & \scalebox{0.8}{0.550} & \scalebox{0.8}{0.304} 
    &\scalebox{0.8}{0.610} &\scalebox{0.8}{0.376} 
    
    &\scalebox{0.8}{0.628} &\scalebox{0.8}{0.379}
    &\scalebox{0.8}{0.626} &\scalebox{0.8}{0.378} 
    &\scalebox{0.8}{0.433} &\scalebox{0.8}{0.295} 
    &\scalebox{0.8}{0.620} &{\scalebox{0.8}{0.336}} 
    & {\scalebox{0.8}{0.414}} &\scalebox{0.8}{0.294} 
    &\scalebox{0.8}{0.804} & \scalebox{0.8}{0.509} \\
    
    \midrule
    \scalebox{0.95}{Weather} 
    & \secondres{\scalebox{0.8}{0.228}} & \secondres{\scalebox{0.8}{0.264}}
    & \boldres{\scalebox{0.8}{0.225}} & \boldres{\scalebox{0.8}{0.257}}
    & {\scalebox{0.8}{0.258}} & {\scalebox{0.8}{0.279}}
    
    & {\scalebox{0.8}{0.238}} & {\scalebox{0.8}{0.268}} 
    & \scalebox{0.8}{0.259} & \scalebox{0.8}{0.315}  
    &\scalebox{0.8}{0.309} &\scalebox{0.8}{0.360} 
    
    &\scalebox{0.8}{0.338} &\scalebox{0.8}{0.382} 
    &\scalebox{0.8}{0.272} &\scalebox{0.8}{0.291} 
    &\scalebox{0.8}{0.246} &\scalebox{0.8}{0.300} 
    &{\scalebox{0.8}{0.259}} &{\scalebox{0.8}{0.287}} 
    &{\scalebox{0.8}{0.237}} &\scalebox{0.8}{0.270} 
    & \scalebox{0.8}{0.292} & \scalebox{0.8}{0.363} \\

    \bottomrule
  \end{NiceTabular}
    \end{small}
  \end{threeparttable}
}
\end{table*}

In this section, we evaluate the efficacy of \abb on time series classification, forecasting, and anomaly detection tasks. 
We show that our \abb can serve as a foundation model with competitive performance on these tasks. The detailed experimental setup is described in Section~\ref{sec:appendix:setup}, while the detailed experimental results are presented in Section~\ref{sec:appendix:res} in the Appendix.

\subsection{Classification}
\paragraph{Datasets.}
We examine the classification ability of \abb on a total of 116 datasets, including 85 uni-variate UCR datasets \cite{dau2018ucr}, 26 multi-variate UEA datasets \cite{bagnall2018uea}. We also include another 5 datasets, i.e., two biomedical datasets, namely, Sleep-EDF dataset \cite{sleepEDF_dataset} for EEG-based sleep stage classification and MIT-BIH dataset \cite{mit_arr_dataset} for ECG-based arrhythmia classification, and three human activity recognition (HAR) datasets, namely, UCIHAR \cite{uciHAR_dataset}, WISDM \cite{wisdm_dataset}, and HHAR \cite{hhar_dataset}.
These datasets have different characteristics and they span a wide range of time series applications. More details about these datasets are included in Appendix~\ref{appen:clf_datasets}.

\paragraph{Baselines and Experimental Settings.}
We select eight state-of-the-art baselines, i.e., GPT4TS \cite{onefitsall}, TimesNet \cite{Timesnet}, ROCKET \cite{ROCKET}, TS-TCC \cite{tstcc}, TS2Vec \cite{ts2vec}, Crossformer~\cite{Crossformer} and PatchTST~\cite{PatchTST} as they showed the best classification accuracy over other Transformer-based architectures. Last, we experiment with a simple single-layer MLP.

\begin{table*}[h]
\caption{Anomaly detection task. We calculate the F1-score (as \%) for each dataset. $\ast$. in the Transformers indicates the name of $\ast$former. \boldres{Blue}: best,  \secondres{Purple}: second best. Table \ref{tbl:anomaly_full} in the Appendix shows the full results.}
\label{tbl:anomaly_short}
\begin{center}
\begin{small}
\scalebox{0.82}{
\setlength\tabcolsep{3pt}
\begin{threeparttable}[b]
\begin{tabular}{c|cccccccccccccccc}
\toprule

\multirow{2}{*}{Methods} 
& \abb & \multirow{2}{*}{GPT4TS} & \multirow{2}{*}{TimesNet} & \multirow{2}{*}{PatchTST} & \multirow{2}{*}{ETS.} & \multirow{2}{*}{FED.} & \multirow{2}{*}{LightTS} & \multirow{2}{*}{DLinear} & \multirow{2}{*}{Stationary} & \multirow{2}{*}{Auto.} & \multirow{2}{*}{Pyra.} & \multirow{2}{*}{Anomaly.} & \multirow{2}{*}{In.} & \multirow{2}{*}{Re.} & \multirow{2}{*}{LogTrans.} & \multirow{2}{*}{Trans.} \\
&\textbf{\textit{(Ours)}}&&&&&&&&&&&&&&& \\

\midrule
SMD & \boldres{87.91} & \secondres{86.89} &84.61 &84.62&83.13& 85.08& 82.53& 77.10 &84.72 &85.11 &83.04& 85.49 &81.65 &75.32& 76.21& 79.56 \\
MSL&\secondres{83.32}&82.45&81.84&78.70&\boldres{85.03}&78.57&78.95&84.88&77.50&79.05&84.86&83.31&84.06&84.40&79.57&78.68\\
SMAP& \boldres{75.96} & \secondres{72.88}&69.39&68.82& 69.50& 70.76& 69.21& 69.26& 71.09& 71.12& 71.09& 71.18& 69.92& 70.40& 69.97& 69.70 \\
SWaT& 92.80 & \boldres{94.23}& \secondres{93.02}&85.72 & 84.91& 93.19 &93.33& 87.52& 79.88& 92.74& 91.78& 83.10& 81.43& 82.80& 80.52& 80.37 \\
PSM& \boldres{97.73} & 97.13&  \secondres{97.34}&96.08& 91.76& 97.23& 97.15& 93.55& 97.29& 93.29& 82.08& 79.40& 77.10& 73.61 &76.74 &76.07 \\
\midrule
Average & \boldres{87.54} & \secondres{86.72}& 85.24&82.79& 82.87& 84.97& 84.23& 82.46& 82.08& 84.26& 82.57 &80.50& 78.83& 77.31& 76.60& 76.88\\

\bottomrule
\end{tabular}

\end{threeparttable}

}
\end{small}
\end{center}

\vskip -0.1in

\end{table*}

\paragraph{Results.}
Table~\ref{tbl:classification_short} reports the classification results, where our proposed \abb demonstrates superior performance over state-of-the-art baselines. Notably, convolution-based methods, including ROCKET, TS-TCC, and our approach, outperform Transformer-based models, highlighting their superiority in classification tasks. For example, in the UCR repository, \abb achieves an impressive accuracy of 83.18\%, outperforming other models including the ROCKET, which scores 81.42\%. The UEA repository results further reinforce our efficacy, with a 72.73\% accuracy, compared to the next best model, PatchTST, at 69.38\%. In more specialized datasets like biomedical signals and HAR, our advantage is even more pronounced, achieving an overall accuracy of 90.24\% and 97.46\%, respectively. These results highlight the robustness and adaptability of \abb in diverse time series contexts.

In our comparative analysis, Transformer models generally face challenges across various datasets, reflecting inherent limitations in handling time series data. MLP models perform well on simpler UCR datasets but falter in complex, noisy environments. TimesNet excels in datasets rich in frequency information but struggles with simpler ones. Last, the GPT4TS model shows promise in larger datasets due to the high capacity of the GPT model, yet underperforms in smaller datasets due to probable overfitting.

\subsection{Forecasting}
\paragraph{Datasets.} 
To assess the efficacy of \abb in forecasting, we conduct comprehensive evaluations on eight benchmark datasets. i.e., Electricity (\textit{ECL}) featuring electricity consumption data, four \textit{ETT} datasets (\textit{ETTh1, ETTh2, ETTm1, ETTm2}) that encompass a range of scenarios in energy transfer technology, \textit{Exchange} that encompasses fluctuating currency exchange rates, \textit{Traffic} that comprises traffic flow information, and \textit{Weather} that offers insights into various meteorological variables over time. We include more details about their characteristics in Appendix~\ref{appen:fore_datasets}.

\paragraph{Baselines and Experimental Settings.}
We compare \abb against a variety of state-of-the-art baselines. For Transformer architectures, we compare against iTransformer \cite{itransformer}, PatchTST, Crossformer, FEDformer \cite{fedformer}, and Autoformer \cite{Autoformer}. For MLP-based models, we compare against RLinear \cite{RLinear} and DLinear \cite{DLinear} models. For general-purpose time series models, we compare our model against TimesNet and GPT4TS. For a convolutional-based forecasting model, we compare with SCINet \cite{SCINet}. Last, we include Time-LLM \cite{timellm}, which is based on Large-Language Models. Similar to \cite{onefitsall} settings, we set the look-back window to 336 for the ETT dataset, 96 for Exchange, 512 for the Traffic and Weather datasets, and 96 for the ECL dataset. We also incorporate the data normalization block, and reverse instance norm in the forecasting task \cite{kim2021reversible}.
For the baselines, we report the best results in their original works if they are consistent with our settings, otherwise, we re-run their codes again.

\paragraph{Results.}
In our forecasting experiments presented in Table~\ref{tbl:forecasting_short}, we notice the superiority of Time-LLM due to its reliance on the large Llama-7B model \cite{touvron2023llama}, which enables it to capture complex patterns and dependencies in data. Other than Time-LLM, \abb consistently outperforms baseline models across various datasets. Specifically, it achieves the second lowest MSE and MAE in seven out of eight datasets, showing 3\% and 3.8\% MSE improvement over the state-of-the-art PatchTST in ETT(avg) and Weather datasets respectively. This indicates the effectiveness of our model in handling datasets with diverse characteristics and complexities. In addition, it shows the effect of the added capability of the ASB module in learning long-range dependencies. 

The results also suggest the superiority of our model over specialized Transformer-based architectures and MLP-based models. These models, e.g., iTransformer and Dlinear show competitive performance in certain datasets but fall behind in others. In addition, GPT4TS shows the power of the GPT models in the forecasting task by scoring the second-best performance in some datasets.

While Time-LLM offers slightly better performance, its computational cost is significantly higher than \abb. To illustrate, \abb demonstrates a nearly equivalent performance to Time-LLM on the ETTh1 dataset with an MSE of 0.413 compared to Time-LLM's 0.408, yet \abb does so with significantly lower computational cost of 6.9e+10 FLOPS against 7.3e+12 for Time-LLM. This showcases the effective balance between performance and computational efficiency in our \abb.

\subsection{Anomaly Detection}
\paragraph{Datasets.}
In this study, we focus on detecting anomalies in unsupervised time series data. We use five benchmark datasets for our experiments: SMD \cite{SMD} for server monitoring, MSL \cite{MSL_SMAP} for space telemetry, SMAP \cite{MSL_SMAP} for earth observations, SWaT \cite{SWaT} for water treatment security, and PSM \cite{PSM} for industrial pump sensors. We discuss their details in Appendix~\ref{appen:anomaly_datasets}.

\paragraph{Baselines and Experimental Settings.}
We followed the same experimental settings and adopted the same baselines in GPT4TS \cite{onefitsall}. These are GPT4TS, TimesNet, PatchTST, ETSformer \cite{woo2022etsformer}, FEDformer, LightTS \cite{lightts}, DLinear, Stationary \cite{Stationary}, Autoformer, Pyraformer \cite{pyraformer}, Anomalyformer \cite{xu2021anomaly}, Informer, Reformer, LogTransformer \cite{LogTransformer}, and the vanilla Transformer. For data preparation, we segmented each dataset with a sliding window, following \cite{xu2021anomaly}. We adopted the reconstruction error as our evaluation metric, common in unsupervised learning for spotting anomalies.

\paragraph{Results.}

Table~\ref{tbl:anomaly_short} presents the results, where \abb performs best in most of the datasets with an overall F1-score of 87.54\%. It outperforms advanced models like FEDformer and Autoformer, especially in the SMD and PSM datasets with F1-scores of 87.91\% and 97.73\% respectively. GPT4TS model follows closely, ranking second with an overall average of 86.72\%. Its high capacity makes it effective in detecting anomalies, though it slightly trails behind.

Notably, Transformer-based models exhibit lower efficacy in anomaly detection in general. This could be regarded to the attention mechanism focusing on dominant normal points, thus missing rare anomalies. Models that consider periodicity, like TimesNet and FEDformer, perform well, indicating the value of periodic analysis in highlighting unusual patterns.

\begin{table}[t]
\centering
\caption{Ablation study to the effect of each component. ASB-L refers to the local filters in the ASB. UWaveGL is the UWaveGestureLibrary dataset from the UEA repository.}
\label{tbl:ablation}
\resizebox{\columnwidth}{!}{
\begin{NiceTabular}{l|cc|cc}
\toprule
\multirow{2}{*}{\textbf{Variant}}& \multicolumn{2}{c|}{\rotatebox{0}{\textbf{Classification (ACC \%)}}} & \multicolumn{2}{c}{\rotatebox{0}{\textbf{Forecasting (MSE)}}} \\
& FordA & UWaveGL & ETTh1 & Exchange \\
\midrule
w/o ASB  & 87.3 & 77.5 & 0.421 & 0.380\\
w/o ASB (L)  & 92.7 & 88.9 & 0.417 & 0.373\\
w/o ICB   & 91.3 & 86.2 & 0.419 & 0.376\\
w/o pretraining & 92.5 & 90.6 & 0.415 & 0.372\\ \midrule
\abb & 93.1 & 91.3 & 0.413 & 0.369 \\ 
\bottomrule
\end{NiceTabular}
}
\end{table}

\section{Model Analysis}
\subsection{Ablation Study}

In Table~\ref{tbl:ablation}, we assess the contribution of the different components in our model, where we report the performance of the model when removing each component individually. Notably, removing the Adaptive Spectral Block (i.e., w/o ASB) yields a notable decline in performance. For classification tasks on FordA and UWaveGestureLibrary datasets, the accuracy drops to 87.3\% and 77.5\%, respectively. Similarly, its absence results in higher MSE values in the forecasting task of 0.421 and 0.380 for ETTh1 and Exchange datasets. This underscores the ASB's critical role in feature extraction and noise reduction. Similarly, excluding the local adaptive part of the ASB (i.e., w/o ASB-L) affects the noisy datasets more than less noisy ones, highlighting the local component’s value in handling noise.

The effect of the ICB was less than the ASB, with less performance degradation in the two tasks. However, its removal shows reduced classification accuracy and increased forecasting MSE indicating its importance. The role of pretraining is similarly validated, as its absence slightly diminishes the model’s performance across both tasks.

\subsection{Efficacy of Adaptive Filtering in Noise Reduction}
\begin{figure}[!tb]
\centering
    \begin{subfigure}[b]{0.23\textwidth}
       \includegraphics[width=\linewidth]{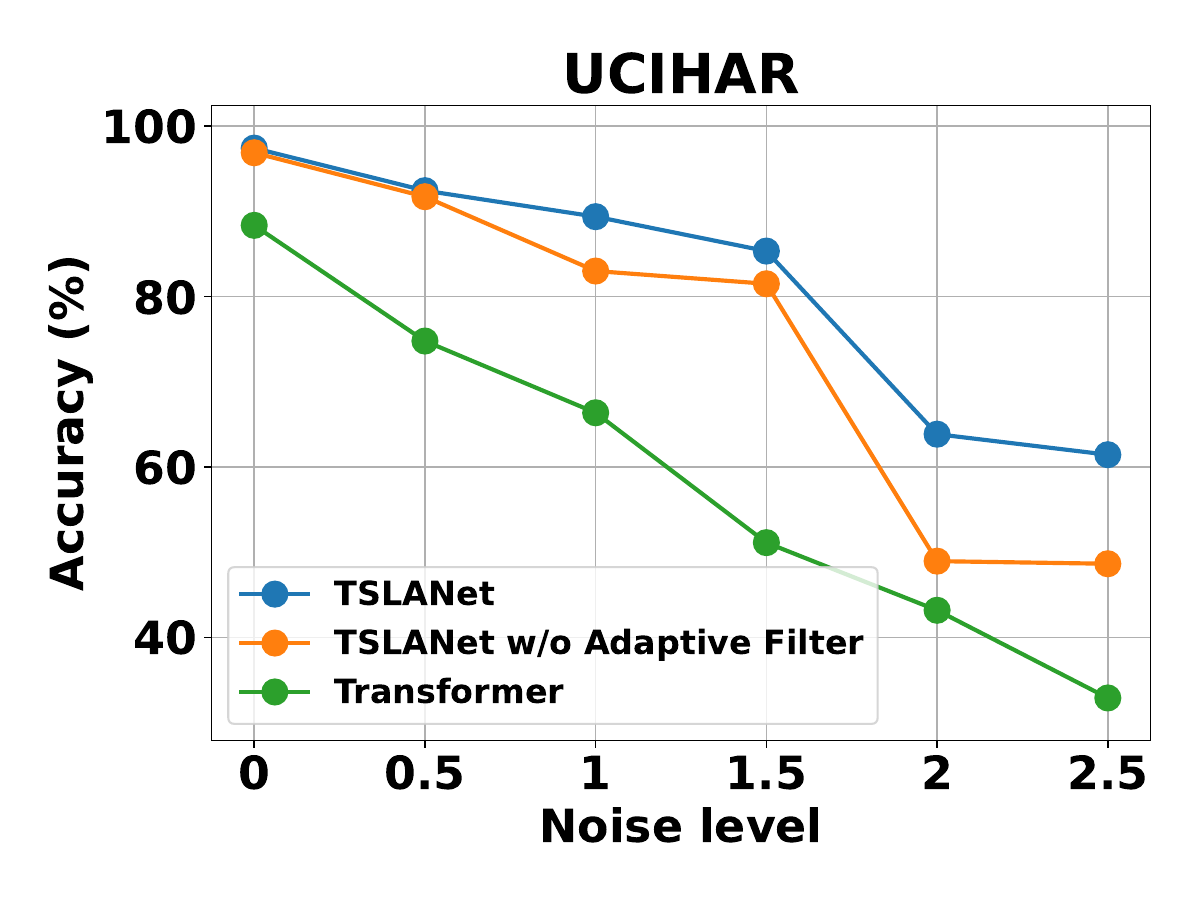} 
        \caption{Robustness against noise levels on UCIHAR dataset.}
        \label{fig:noise_levels_part1}
    \end{subfigure}
    ~ 
    \begin{subfigure}[b]{0.23\textwidth}
       \includegraphics[width=\linewidth]{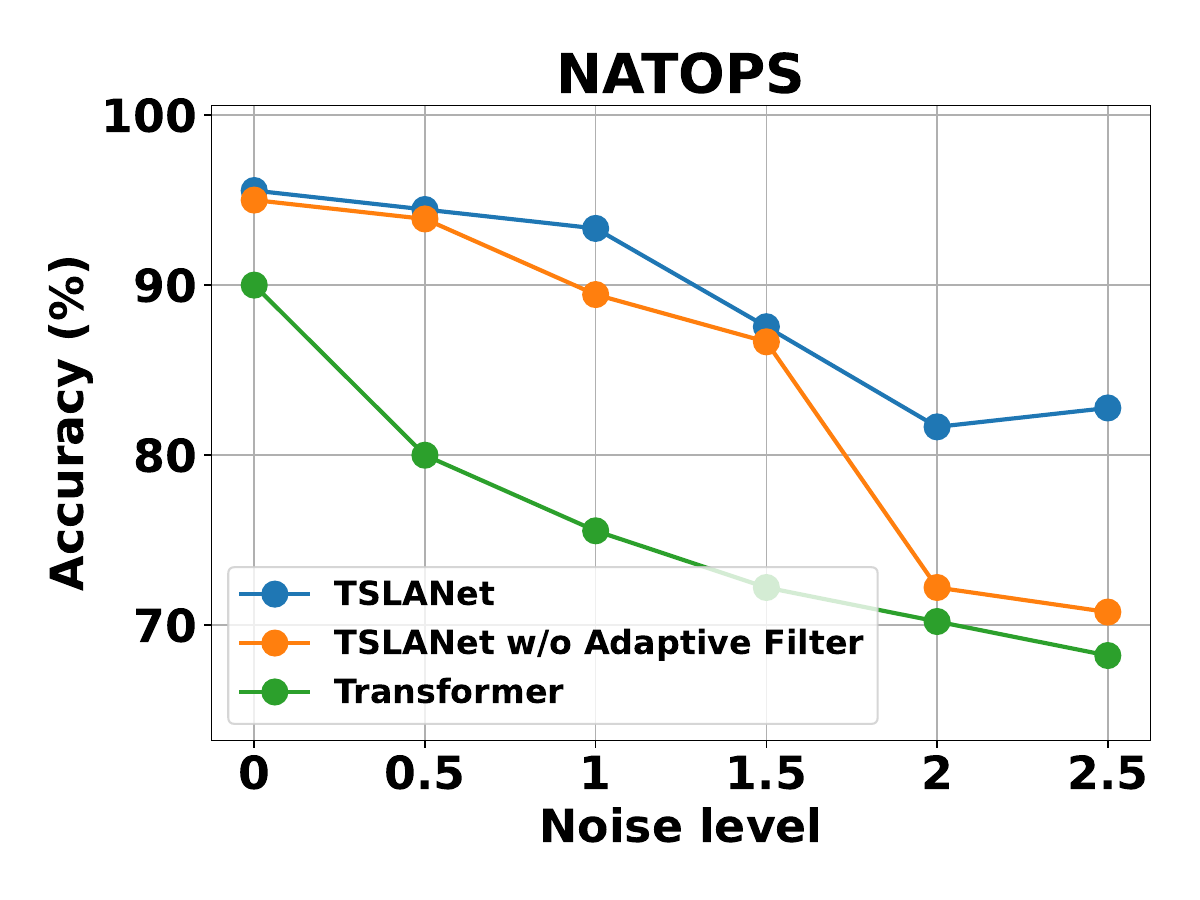} 
        \caption{Robustness against noise levels on NATOPS dataset.}
        \label{fig:noise_levels_part2}
    \end{subfigure}
    \begin{subfigure}[b]{0.47\textwidth}
       \centering
        \includegraphics[width=\columnwidth]{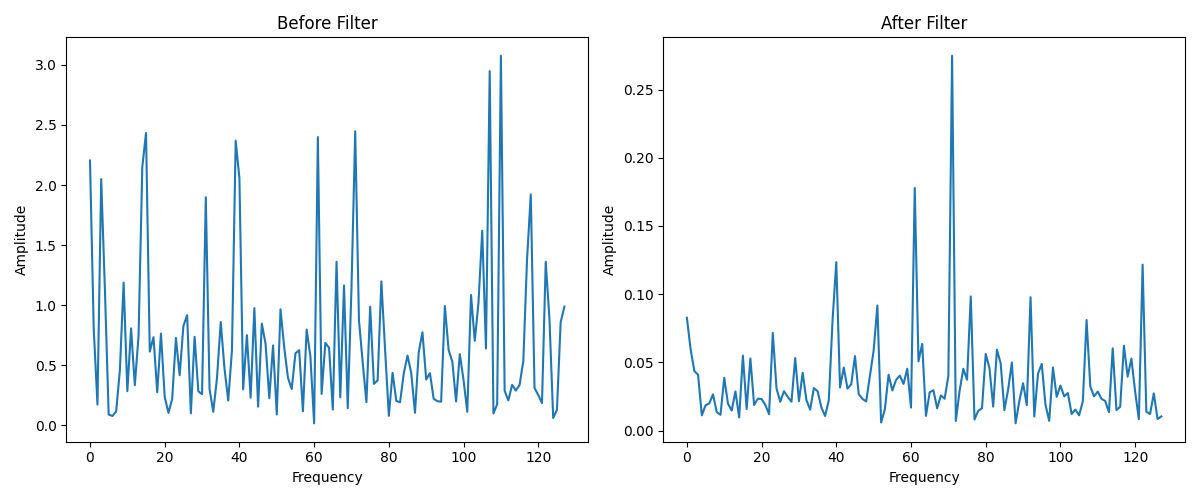}
        \caption{The features before and after the local adaptive filter on the UEA Handwriting dataset.}
        \captionsetup{skip=2mm}
        \label{fig:adaptive_filter_effect}
    \end{subfigure}
\caption{Effectiveness of the Adaptive Filter in noise reduction.}
\label{fig:adaptive_filter_effectiveness}
\end{figure}

We delve into the effectiveness of the Adaptive Filter in mitigating noise and enhancing model robustness by examining Figure~\ref{fig:adaptive_filter_effectiveness}. Specifically, Figures~\ref{fig:noise_levels_part1} and~\ref{fig:noise_levels_part2} present the performance of \abb, both with and without the Adaptive Filter, against the Transformer model by adding different Gaussian noise levels to the time series.
The performance of the Transformer deteriorates rapidly as noise increases. In contrast, \abb maintains a relatively stable performance, with the variant using the Adaptive Filter showing the most resilience to noise. This is particularly noteworthy at higher noise levels, where the accuracy of the standard Transformer falls steeply, while \abb with the Adaptive Filter experiences a much less pronounced decline.

In Figure~\ref{fig:adaptive_filter_effect}, we observe the frequency spectra before and after applying the Adaptive Filter. The left plot shows a noisy spectrum with high amplitude spikes across various frequencies. However, after applying the Adaptive Filter, a markedly cleaner spectrum where the amplitude of noise spikes is significantly reduced, particularly in the higher frequency range. This demonstrates the filter's ability to attenuate unwanted noise while preserving the relevant signal.

\subsection{Scaling Efficiency}

\begin{figure}[!tb]
    \centering
    \includegraphics[width=\columnwidth]{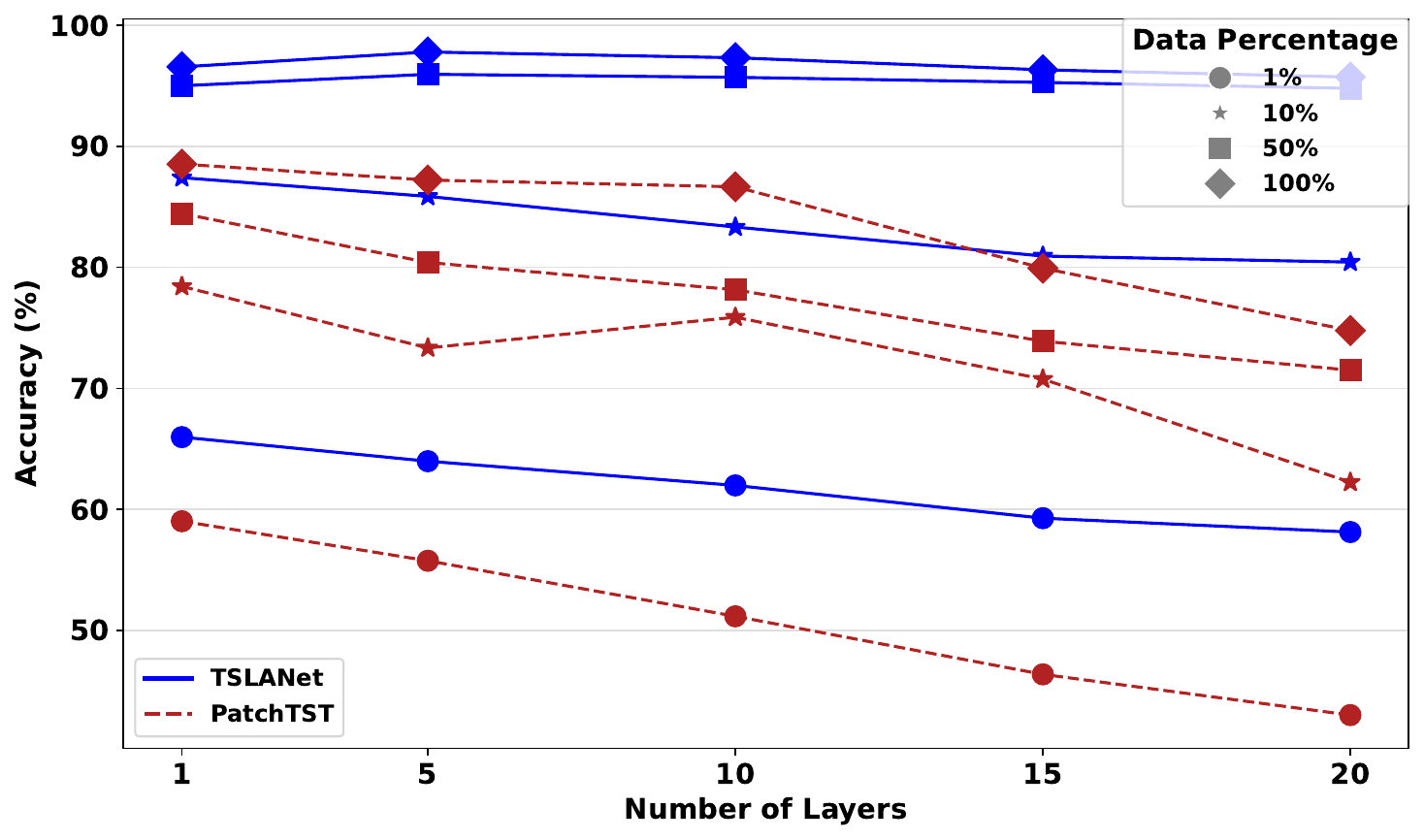}
    \captionsetup{skip=2mm}
    \caption{A comparison between \abb vs. PatchTST in terms of accuracy with varying the number of layers in both for different data percentages from the uWaveGestureLibraryAll dataset.}
    \label{fig:complexity_analysis}
\end{figure}

We compare the scalability of our \abb with one of the best-performing Transformer models in the classification task, i.e., PatchTST \cite{PatchTST}, by observing their performance across various dataset sizes and layer counts. Specifically, we experiment with variable data sizes from the uWaveGestureLibraryAll dataset, as shown in Figure~\ref{fig:complexity_analysis}. Notably, in smaller data sizes, \abb demonstrates a consistent accuracy level, subtly decreasing as the number of layers increases. In contrast, the PatchTST shows a marked decline in accuracy with additional layers, suggesting a potential overfitting issue or inefficiency in handling limited data with increased model complexity.

As dataset sizes grow, \abb performance remains robust, showing slight variations in accuracy with more layers. This stability contrasts with the PatchTST performance, which tends to decrease notably at higher layer counts. This trend in PatchTST could be attributed to their inherent design, which might lead to diminishing returns or optimization challenges as the model depth increases. Lastly, we notice that \abb effectively leverages larger dataset samples, as its performance improves with an increase in the number of layers, highlighting its capacity to capitalize on more extensive data for enhanced accuracy.

\subsection{Complexity Analysis}
\label{sec:complexity_analysis}
\begin{figure}[!tb]
    \centering
    \includegraphics[width=\columnwidth]{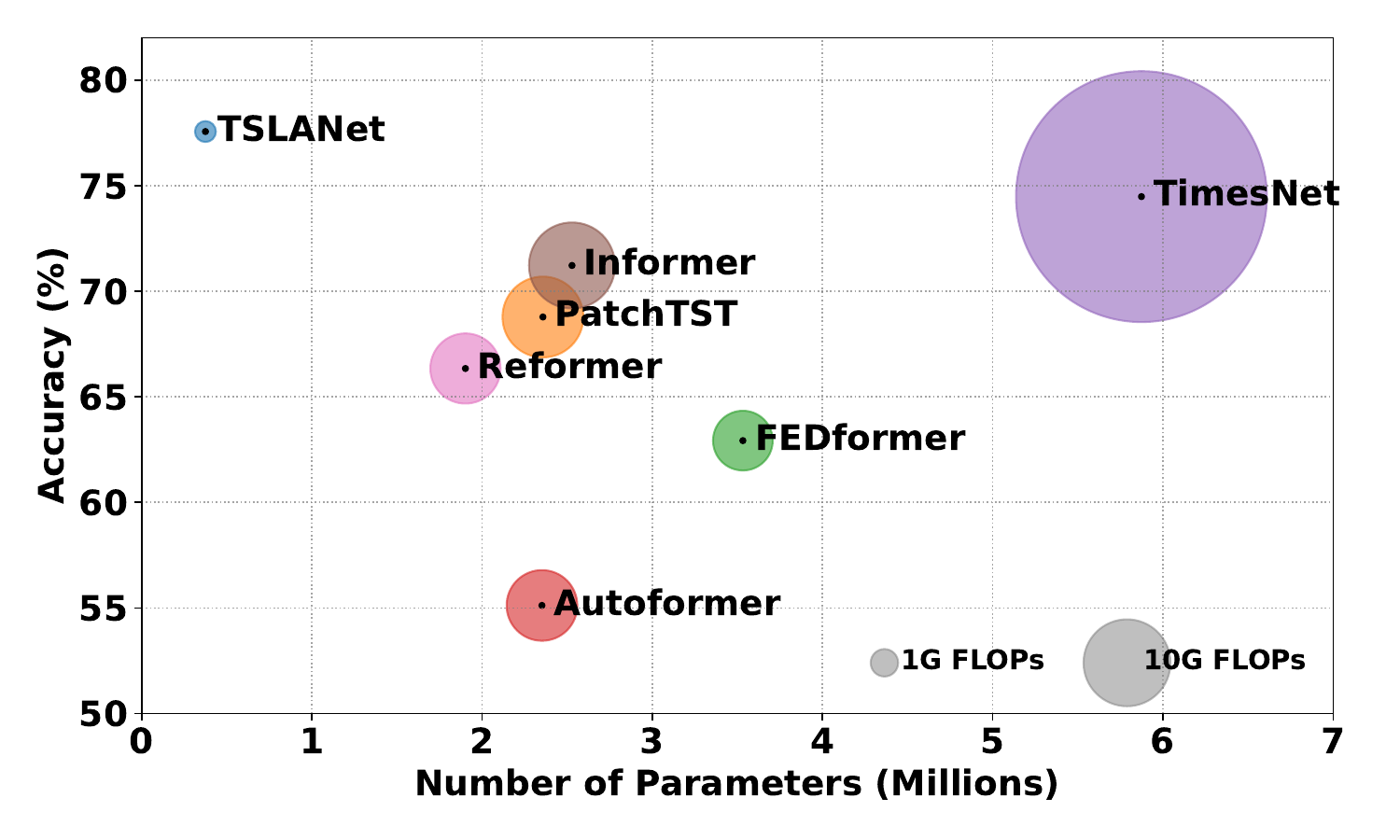}
    \caption{\abb vs. baselines in terms of the number of parameters and FLOPS count against the classification accuracy of the UEA Heartbeat dataset.}
    \label{fig:param_flops_acc}
\end{figure}

We compare the complexity of our \abb with TimesNet and Transformer-based models, e.g., PatchTST, FEDFormer, AutoFormer, Informer, and Reformer in terms of the number of parameters, FLOPs, and accuracy on the UEA Heartbeat dataset, as shown in Figure~\ref{fig:param_flops_acc}.
\abb demonstrates superior efficiency and accuracy in time series analysis, achieving the highest accuracy of 77.56\% with the lowest computational and parameter footprint among the compared models. It requires 93\% fewer FLOPs and 84\% fewer parameters than the PatchTST, yet outperforms it by over 8\% in accuracy. Compared to TimesNet, TSLANet operates with more than 99\% fewer FLOPs and parameters while still delivering a 3\% higher accuracy.

This considerable reduction in computational demand confirms the lightweight nature of \abb compared to Transformer-based alternatives, underscoring its capacity to make time series analysis more efficient.

\section{Conclusions}
In this paper, we introduced \abb, a novel lightweight model for time series analysis that revisits the convolution approach as a potent replacement to Transformers, with an innovative combination of convolution operations and adaptive spectral analysis. Our comprehensive experiments across various datasets in classification, forecasting, and anomaly detection have demonstrated its superior performance over traditional Transformer models, particularly in its ability to maintain high accuracy levels in noisy conditions and across different data sizes. Furthermore, our in-depth layer-wise performance analysis revealed that \abb not only outperforms Transformers in smaller datasets but also exhibits improved scalability with increasing layers, particularly in larger datasets. \abb is a step towards a foundation model for time series analysis.

\section*{Impact Statement}
Our proposed work \abb aims to advance the field of Machine Learning by providing a more efficient, scalable, and robust foundation model for analyzing time series data across various applications. It has the potential to impact various sectors, including healthcare, finance, and environmental monitoring, by enhancing forecasting accuracy and anomaly detection capabilities. Such improvements could lead to better patient outcomes, more informed financial decisions, and greater preparedness for natural disasters.

\bibliography{ref}
\bibliographystyle{icml2024}

\newpage
\appendix
\onecolumn  

\section{Circular Convolutions}
\label{sec:appen1:cir_conv}
The convolution theorem suggests that the multiplication in the frequency domain is equivalent to the circular convolution process.

Let \( x[n] \) and \( h[n] \) be two length \( N \) sequences. Their DFTs are \( X[k] \) and \( H[k] \), respectively. Consider the circular convolution \( y[n] = (x \circledast h)[n] \). The DFT of \( y[n] \) is \( Y[k] \).

First, the DFT of the Convolution can be formulated as:
   \[ Y[k] = \sum_{n=0}^{N-1} \left( \sum_{m=0}^{N-1} x[m] \cdot h[(n-m) \mod N] \right) \cdot e^{-i 2 \pi k n / N} \]

However, if we changed the order of summation, it becomes:
   \[ Y[k] = \sum_{m=0}^{N-1} x[m] \cdot \sum_{n=0}^{N-1} h[(n-m) \mod N] \cdot e^{-i 2 \pi k n / N} \]

By substituting \( n-m \) with \( r \):
   \[ Y[k] = \sum_{m=0}^{N-1} x[m] \cdot e^{-i 2 \pi k m / N} \cdot \sum_{r=0}^{N-1} h[r] \cdot e^{-i 2 \pi k r / N} \]

Therefore, we recognize the DFTs of \( x[n] \) and \( h[n] \):
   \[ Y[k] = \left( \sum_{m=0}^{N-1} x[m] \cdot e^{-i 2 \pi k m / N} \right) \cdot \left( \sum_{r=0}^{N-1} h[r] \cdot e^{-i 2 \pi k r / N} \right) \]
   \[ Y[k] = X[k] \cdot H[k] \]

Thus, we have shown that the DFT of the circular convolution of two sequences \( x[n] \) and \( h[n] \) is the product of their individual DFTs, i.e., \( Y[k] = X[k] \cdot H[k] \).

\section{Frequency Domain Processing Role to Learn Long-Range Dependencies}
Fourier transforms, used in our Adaptive Spectral Block (ASB), can learn long-range and short-range dependencies in time series. The Fourier Transform (FT) of a time series \(x(t)\) is given by:
   \[X(f) = \int_{-\infty}^{\infty} x(t) e^{-j2\pi ft} dt\]
   where \(X(f)\) represents the signal in the frequency domain, \(f\) is the frequency, and \(t\) represents time.

The FT decomposes \(x(t)\) into its constituent frequencies, where each frequency component represents a pattern in the time series. Low-frequency components correspond to long-range dependencies (slowly changing trends), and high-frequency components correspond to short-range dependencies (rapid fluctuations). Let's consider a simplified model where the ASB applies a filter \(H(f)\) to the Fourier transform \(X(f)\) of the input signal, enhancing certain frequencies while attenuating others: \[Y(f) = H(f) \cdot X(f)\] where \(Y(f)\) is the output signal in the frequency domain.

The adaptiveness comes from adjusting \(H(f)\) based on the data, which can be modeled as a learning process where \(H(f)\) is updated to minimize a loss function \(L\) that measures the discrepancy between the model output and the true data characteristics:
   \[\min_{H(f)} L(Y(f), \text{True Data})\]
Through this process, \(H(f)\) learns to emphasize the frequency components that are most relevant for predicting the target, whether they capture long-range or short-range dependencies.

After filtering in the frequency domain, the inverse Fourier transform (IFT) is applied to convert \(Y(f)\) back into the time domain, yielding the modified signal \(y(t)\):
   \[y(t) = \int_{-\infty}^{\infty} Y(f) e^{j2\pi ft} df\]
This signal now encapsulates the learned dependencies, ready for further processing or as an input to subsequent model layers.

\section{Algorithm of Adaptive Spectral Block}
\begin{algorithm}[!h]
\caption{Pseudocode of the Adaptive Spectral Block.}
\label{alg:adaptive_global_filter}
\definecolor{codeblue}{rgb}{0.25,0.5,0.5}
\definecolor{codekeyword}{rgb}{0.13,0.13,1} 
\lstset{
  backgroundcolor=\color{white},
  basicstyle=\fontsize{7.2pt}{7.2pt}\ttfamily\selectfont,
  columns=fullflexible,
  breaklines=true,
  captionpos=b,
  commentstyle=\fontsize{7.2pt}{7.2pt}\color{codeblue},
  keywordstyle=\fontsize{7.2pt}{7.2pt}\color{codekeyword}, 
  language=Python,
  morekeywords={def, in}, 
}
\begin{lstlisting}[language=python]
def adaptive_high_freq_mask(x, threshold):
    # Calculate energy
    energy = torch.abs(x_fft).pow(2).sum(dim=-1)

    # Compute the adaptive threshold
    threshold = torch.quantile(energy, threshold)
    
    # Identify the dominant frequencies
    dominant_freq = normalized_energy > threshold
    
    # Set adaptive mask values
    adaptive_mask[dominant_freq] = 1
    
    return adaptive_mask

# Transform input x_in to frequency domain 
X_fft = fft(x_in)

# Create an adaptive mask for high-freq. components
freq_mask = adaptive_high_freq_mask(X_fft, threshold)

# Apply adaptive high-frequency mask
X_masked = X_fft * freq_mask

# Apply global and local learnable weights
X_L = X_masked * local_weight
X_G = X_fft * global_weight

# Transform data back into the time domain
x_out = ifft(X_L + X_G)

\end{lstlisting}
\end{algorithm}

\section{Experimental Setup}
\label{sec:appendix:setup}
\subsection{Training Protocol}
To train the classification experiments, we optimized \abb using AdamW with a learning rate of 1e-3 and a weight decay of 1e-4, applied during both training and pretraining phases. The experiments ran for 50 epochs for pretraining and 100 epochs for fine-tuning. For the forecasting and anomaly detection experiments, we utilized a learning rate of 1e-4 and a weight decay of 1e-6, with both phases running for 10 and 20 epochs.

For all experiments, the stride was set to half of the patch size to ensure overlapping windows. Each experiment was repeated three times, with the average performance reported. \abb was implemented using PyTorch and conducted on NVIDIA RTX A6000 GPUs.

\subsection{Objective Functions}
For the classification task, we employ a categorical cross-entropy loss function with label smoothing, defined as \( \mathcal{L}_{\text{clf}} = -\sum_{i=1}^{C} y_{i}^{\text{smooth}} \cdot \log(\hat{y}_{i}) \). Here, \( y_i^{\text{smooth}} \) is the true class label in one-hot encoded form adjusted via label smoothing, \( \hat{y}_{i} \) is the predicted probability for each class, and \( C \) is the total number of classes. Label smoothing reduces model confidence by adjusting the true labels with a smoothing parameter \( \epsilon \), making the distribution more uniform, where each \( y_i \) is transformed to \( y_{i}^{\text{smooth}} = (1 - \epsilon) \cdot y_{i} + \frac{\epsilon}{C} \).

In forecasting and anomaly detection, we use the Mean Squared Error (MSE) to measure discrepancies between predicted values and actual observations, expressed as \( \mathcal{L}_{\text{MSE}} = \frac{1}{N}\sum_{i=1}^{N} (y_i - \hat{y}_i)^2 \). Here, \( y_i \) represents the actual value at time \( i \), \( \hat{y}_i \) denotes the forecasted value, and \( N \) is the number of predictions. This MSE loss is also utilized in self-supervised learning tasks to reconstruct masked patches.

\subsection{Evaluation Metrics}
Model performance was evaluated using standard metrics appropriate to each task. For classification, we reported accuracy; for forecasting, Mean Squared Error (MSE) and Mean Absolute Error (MAE) were used; for anomaly detection, the F1-score was our primary metric due to the imbalanced nature of the datasets.

\section{Datasets Details}
\subsection{Data Preprocessing}
For the classification task, the UCR and UEA datasets are already split into train/test splits. A validation set was picked from each dataset in the training set with a ratio of 80/20. The selection of the hyperparameters was based on the average results on the validation sets across each collection of datasets, i.e., UCR and UEA. 
For biomedical and human activity recognition datasets, which are not split by default, we split the data into a 60/20/20 ratio for train/validation/test splits.
For forecasting and anomaly detection datasets, these are split into a ratio of 70/10/20 following a line of previous works, 
towards a fair comparison with these works \cite{fedformer,reformer,Informer,Timesnet}.
All datasets are normalized during training.

For the self-supervised task, we deploy the unlabeled version of the training set in each dataset for pretraining, then use the same set again with labels for fine-tuning.

\subsection{Classification}
\label{appen:clf_datasets}
In our evaluation, we extensively utilize four categories of datasets:
\begin{itemize}
    \item \textbf{UCR datasets:} The UCR Time Series Classification Archive is one of the most comprehensive collections of univariate datasets tailored for time series analysis. This archive encompasses 85 diverse datasets, each presenting unique challenges and characteristics that span a wide array of domains, from healthcare and finance to environmental monitoring and beyond. The variety within the UCR archive allows for a robust assessment of \abb across different contexts, showcasing its versatility and performance. 

    \item \textbf{UEA datasets:} We also incorporate datasets from the University of East Anglia (UEA) Time Series Classification repository, which is renowned for its rich collection of multivariate time series datasets. We were able to preprocess 26 datasets, each offering a multidimensional perspective on time series analysis across various real-world scenarios, such as human activity recognition, sensor data interpretation, and complex system monitoring. More details about the UCR and UEA datasets can be found in \url{https://www.timeseriesclassification.com/}.

    \item \textbf{Biomedical datasets:} The biomedical domain presents unique challenges and opportunities for time series analysis. In this context, we utilized two pivotal datasets for our evaluation: the Sleep-EDF dataset and the MIT-BIH Arrhythmia dataset.
    \begin{itemize}
        \item \textbf{Sleep-EDF Dataset:} This dataset consists of polysomnography recordings intended for sleep stage classification. It is part of the PhysioNet database and includes polysomnographic sleep recordings that have been widely used to analyze sleep patterns and stages. For our analysis, we extracted the brain EEG signals.

        \item \textbf{MIT-BIH Arrhythmia Dataset:} Another significant dataset from PhysioNet, the MIT-BIH Arrhythmia Dataset, is composed of electrocardiogram (ECG) recordings used primarily for arrhythmia detection and classification. It is one of the most extensively used datasets for validating arrhythmia detection algorithms, offering a comprehensive collection of annotated heartbeats and arrhythmia examples.
    \end{itemize}
    A summary of the characteristics of these two datasets is presented in Table~\ref{tbl:biomedical_data}.

    \item \textbf{Human Activity Recognition datasets:} Human activity recognition (HAR) using sensor data is a vital application of time series analysis, with implications for health monitoring, elder care, and fitness tracking. In this study, we evaluate our model using three prominent HAR datasets: UCIHAR, WISDM, and HHAR.
    \begin{itemize}
        \item \textbf{UCI Human Activity Recognition Using Smartphones (UCIHAR):} This dataset is collected from experiments that were carried out with a group of 30 volunteers performing six activities (walking, walking upstairs, walking downstairs, sitting, standing, and laying) while wearing a smartphone on the waist. The smartphone's embedded accelerometer and gyroscope captured 3-axial linear acceleration and 3-axial angular velocity, respectively. 

        \item \textbf{Wireless Sensor Data Mining (WISDM):} The WISDM dataset includes time series data from smartphone sensors and wearable devices, capturing various human activities such as walking, jogging, sitting, and standing. It provides a diverse set of user-generated activity data, making it suitable for testing the robustness of HAR models across different motion patterns and sensor placements.

        \item \textbf{Heterogeneity Human Activity Recognition (HHAR):} HHAR dataset stands out due to its collection from multiple device types, including smartphones and smartwatches, across different individuals performing activities like biking, sitting, standing, walking, stair climbing, and more. Its heterogeneity in terms of device types and positions offers a challenging benchmark for assessing a model's ability to generalize across various sensor configurations and activity types. Here, we utilized the data from the Samsung devices.
    \end{itemize}
    A summary of the characteristics of these three datasets is presented in Table~\ref{tbl:har_data}.    
\end{itemize}
\begin{table}[!bth]
\centering
\caption{A description of characteristics of the biomedical datasets used in our experiments.}
    \begin{tabular}{@{}l|ccccc@{}}
    \toprule
    Dataset & \# Train & \# Test & Length & \# Channel & \# Class \\ \midrule
    Sleep EEG & 25,612 & 8,910 & 3,000 & 2 & 5 \\
    Arrhythmia ECG & 70,043 & 21,892 & 187 & 1 & 2 \\
    
    \bottomrule
    \end{tabular}
\label{tbl:biomedical_data}
\end{table}

\begin{table}[!bth]
\centering
\caption{A description of characteristics of the Human Activity Recognition datasets used in our experiments.}
    \begin{tabular}{@{}l|ccccc@{}}
    \toprule
    Dataset & \# Train & \# Test & Length & \# Channel & \# Class \\ \midrule
    UCIHAR & 7,352 & 2,947 & 128 & 9 & 6 \\
    WISDM & 4,731 & 2,561 & 128 & 3 & 6 \\
    HHAR & 10,336 & 4,436 & 128 & 3 & 6 \\
    \bottomrule
    \end{tabular}
\label{tbl:har_data}
\end{table}

\subsection{Forecasting}
\label{appen:fore_datasets}
Our study leverages a diverse set of forecasting datasets to evaluate the effectiveness of our model across various domains:

\begin{itemize}
    \item \textbf{Electricity:} This dataset contains electricity consumption records from 321 clients, offering insights into usage patterns and enabling demand forecasting, crucial for optimizing power generation and distribution.

    \item \textbf{ETT (Electricity Transformer Temperature) datasets:} The ETTh1, ETTh2, ETTm1, and ETTm2 datasets provide data on the temperature of electricity transformers and the load, facilitating the prediction of future temperatures and loads based on past patterns. These datasets vary in granularity, with "h" indicating hourly data and "m" indicating 15-minute intervals, offering a range of temporal resolutions for forecasting challenges.

    \item \textbf{Exchange Rate:} Featuring daily exchange rates of different currencies against the US dollar, this dataset is vital for financial forecasting, enabling models to anticipate currency fluctuations based on historical data.

    \item \textbf{Traffic:} Traffic dataset consists of hourly interstate 94 Westbound traffic volume for the Twin Cities (Minneapolis-St. Paul) metropolitan area, allowing for the prediction of traffic flow patterns, essential for urban planning and congestion management.
    
    \item \textbf{Weather:} This dataset includes hourly weather conditions and atmospheric measurements from a weather station, supporting forecasts of various weather phenomena, crucial for agriculture, transportation, and daily life planning.
\end{itemize}
We describe the characteristics of these datasets in Table~\ref{tbl:forecating_datasets}.
\begin{table}[!h]
  \caption{Descriptions of the forecasting datasets. \emph{Dim} shows the variate number of each dataset. \emph{Dataset Size} indicates the size of the (Train, Validation, Test) split respectively. \emph{Frequency} denotes the sampling interval of time points.}
  \centering
  \begin{threeparttable}
  \begin{small}
  \renewcommand{\multirowsetup}{\centering}
  \setlength{\tabcolsep}{6.5pt}
  \begin{tabular}{l|c|c|c|c}
    \toprule
    Dataset & Dim & Dataset Size & Frequency& Information \\
    \toprule
     ECL & 321 & (18317, 2633, 5261) & Hourly & Electricity \\ \midrule
     ETTh1, ETTh2 & 7 & (8545, 2881, 2881) & Hourly & Electricity\\
     \midrule
     ETTm1, ETTm2 & 7 & (34465, 11521, 11521) & 15min & Electricity\\ \midrule
    Exchange & 8 & (5120, 665, 1422) & Daily & Economy \\ \midrule
    Traffic & 862 & (12185, 1757, 3509) & Hourly & Transportation \\ \midrule
    Weather & 21  & (36792, 5271, 10540) & 10min & Weather\\
    \bottomrule
    \end{tabular}
    \end{small}
  \end{threeparttable}
  \label{tbl:forecating_datasets}
\end{table}

\subsection{Anomaly Detection}
\label{appen:anomaly_datasets}
Anomaly detection plays a pivotal role across various domains, enabling the identification of unusual patterns that may indicate critical incidents, such as system failures, security breaches, or environmental changes. In our study, we assess the performance of our model using five benchmark datasets, each representing a distinct application area, to demonstrate its effectiveness in detecting anomalies in diverse settings:

\begin{itemize}
    \item \textbf{SMD (Server Machine Dataset):} Utilized for server monitoring, the SMD dataset comprises multivariate time series data collected from servers and aims to identify unusual server behaviors that could indicate failures or security issues.

    \item \textbf{MSL (Mars Science Laboratory):} This dataset contains telemetry data from the Mars Science Laboratory rover, focusing on space exploration applications. Anomaly detection in this context is crucial for identifying potential issues with spacecraft systems based on their operational data.

    \item \textbf{SMAP (Soil Moisture Active Passive):} Related to earth observations, the SMAP dataset includes soil moisture measurements intended for environmental monitoring. Detecting anomalies in soil moisture can provide insights into environmental conditions and potential agricultural impacts.

    \item \textbf{SWaT (Secure Water Treatment):} In the domain of water treatment security, the SWaT dataset consists of data from a water treatment testbed, simulating the operational data of water treatment plants. Anomaly detection here is vital for ensuring the safety and security of water treatment processes.

    \item \textbf{PSM (Pump Sensor Monitoring):} Focused on industrial pump sensors, the PSM dataset gathers sensor data from pumps in industrial settings. Anomalies in this dataset can indicate equipment malfunctions or the need for maintenance, critical for preventing industrial accidents.
\end{itemize}
The detailed characteristics of these datasets is presented in Table~\ref{tbl:anomaly_datasets}.

\begin{table}[!h]
  \caption{Descriptions of the Anomaly detection datasets. \emph{Dim} shows the variate number of each dataset. \emph{Dataset Size} indicates the size of the (Train, Validation, Test) split respectively. \emph{Frequency} denotes the sampling interval of time points.}
  \centering
  \begin{threeparttable}
  \begin{small}
  \renewcommand{\multirowsetup}{\centering}
  \setlength{\tabcolsep}{6.5pt}
  \begin{tabular}{l|c|c|c|c}
    \toprule
    Dataset & Dim & Length & Dataset Size & Information \\
    \toprule
        
     SMD & 38 & 100 & (566724, 141681, 708420) & Server Machine \\ \midrule
     MSL & 55 & 100 & (44653, 11664, 73729) & Spacecraft \\
     \midrule
     SMAP & 25 & 100 & (108146, 27037, 427617) & Spacecraft \\ \midrule
    SWaT & 51 & 100 & (396000, 99000, 449919) & Infrastructure \\ \midrule
    PSM & 25 & 100 & (105984, 26497, 87841) & Server Machine \\ 
    \bottomrule
    \end{tabular}
    \end{small}
  \end{threeparttable}
  \label{tbl:anomaly_datasets}
\end{table}

\section{Full Results}
\label{sec:appendix:res}
\subsection{Classification}

{\small
\begin{longtable}{l|c|c|c|c|c|c|c|c|c}
    \toprule
        Dataset & \abb & GPT4TS & TimesNet & ROCKET & CrossF. & Pat.TST & MLP & TS-TCC & TS2VEC \\ \midrule
Adiac  & \boldres{80.56} &  52.69  &  24.04  & \secondres{78.52} &  58.31  &  34.78  &  61.38  &  76.57  &  72.89  \\
ArrowHead  & \boldres{80.57} &  66.29  &  49.71  &  77.31  &  73.71  &  72.57  &  75.43  &  62.20  & \secondres{77.71} \\
Beef  & \boldres{90.00} &  66.67  &  60.00  &  67.33  &  73.33  & \secondres{76.67} &  73.33  &  47.32  &  76.67  \\
BeetleFly  & \boldres{90.00} &  85.00  &  80.00  & \secondres{88.00} &  85.00  &  80.00  &  80.00  &  31.25  &  85.00  \\
BirdChicken  & \boldres{100.00} & \secondres{85.00} &  60.00  &  84.50  &  85.00  &  80.00  &  75.00  &  75.00  &  80.00  \\
CBF  & \boldres{97.56} &  92.00  & \secondres{92.22} &  89.67  &  89.56  &  85.11  &  83.44  &  90.79  &  88.33  \\
Car  &  88.33  &  76.67  &  30.00  & \boldres{99.52} &  86.67  &  75.00  &  86.67  &  71.88  & \secondres{99.22} \\
ChlorineConcentration  & \boldres{85.94} &  61.25  &  55.21  &  69.40  &  61.72  &  56.56  &  61.72  &  57.40  & \secondres{71.85} \\
CinC\_ECG\_torso  & \secondres{85.51} &  23.99  &  51.74  &  84.96  &  84.93  &  66.88  &  46.67  & \boldres{95.55} &  79.28  \\
Coffee  & \boldres{100.00} & \secondres{100.00} &  53.57  &  100.00  &  100.00  &  100.00  &  100.00  &  95.83  &  100.00  \\
Computers  & \secondres{68.40} &  52.00  &  62.40  &  66.48  &  63.20  & \boldres{69.60} &  58.00  &  61.95  &  60.40  \\
Cricket\_X  &  76.15  &  6.41  &  55.64  & \boldres{77.31} &  41.79  &  45.38  &  32.56  & \secondres{77.25} &  76.15  \\
Cricket\_Y  & \secondres{78.72} &  49.74  &  55.90  & \boldres{79.15} &  47.69  &  43.59  &  42.31  &  75.75  &  73.08  \\
Cricket\_Z  & \boldres{80.00} &  8.21  &  57.44  & \secondres{79.33} &  41.79  &  47.95  &  32.56  &  75.83  &  76.92  \\
DiatomSizeReduction  &  92.16  &  88.89  &  48.69  & \secondres{97.68} &  95.75  &  91.18  &  93.14  &  95.94  & \boldres{97.71} \\
DistalPhalanxOutlineAgeGroup  & \boldres{86.50} & \secondres{86.50} &  80.25  &  76.12  &  80.75  &  82.00  &  80.50  &  85.25  &  81.25  \\
DistalPhalanxOutlineCorrect  &  80.67  &  75.67  &  73.67  &  75.68  &  76.17  &  78.50  &  75.50  & \secondres{80.76} & \boldres{81.17} \\
DistalPhalanxTW  & \boldres{80.50} &  78.25  &  77.25  &  70.07  &  79.25  &  79.00  & \secondres{79.50} &  79.50  &  78.00  \\
Earthquakes  & \boldres{82.30} &  38.82  &  23.60  &  75.32  & \secondres{82.30} &  80.75  &  59.94  &  75.89  &  72.36  \\
ECG200  &  88.00  &  85.00  & \boldres{90.00} &  84.90  &  86.00  & \secondres{89.00} &  84.00  &  87.50  &  87.00  \\
ECG5000  & \secondres{94.62} &  93.40  &  93.47  & \boldres{94.72} &  94.36  &  93.87  &  94.18  &  94.19  &  93.33  \\
ECGFiveDays  &  99.30  &  94.77  &  83.74  & \boldres{100.00} &  98.49  &  86.41  &  96.63  &  90.71  & \secondres{100.00} \\
ElectricDevices  &  68.28  &  56.36  &  68.58  &  66.84  &  61.87  & \boldres{74.66} &  48.22  & \secondres{69.31} &  68.10  \\
FaceAll  &  82.31  &  37.22  &  73.61  & \boldres{93.33} & \secondres{90.53} &  79.94  &  78.64  &  76.99  &  79.17  \\
FaceFour  & \boldres{94.32} &  7.95  &  52.27  &  77.39  &  93.18  &  86.36  &  82.95  &  85.42  & \secondres{94.32} \\
FacesUCR  &  92.39  &  82.88  &  46.00  & \boldres{94.81} &  83.07  &  77.46  &  74.39  &  92.93  & \secondres{94.24} \\
FiftyWords  & \boldres{80.00} &  36.48  &  61.32  &  76.92  &  62.86  &  55.16  &  58.90  &  77.62  & \secondres{79.12} \\
FISH  & \secondres{94.29} &  71.43  &  59.43  & \boldres{96.86} &  84.57  &  71.43  &  87.43  &  61.29  &  93.14  \\
FordA  & \boldres{93.06} &  50.49  &  66.20  &  90.61  &  70.62  &  50.90  &  51.32  & \secondres{92.35} &  89.28  \\
FordB  & \secondres{91.39} &  61.99  &  54.43  &  77.53  &  52.70  &  52.20  &  51.16  & \boldres{91.72} &  83.50  \\
Gun\_Point  & \boldres{99.33} &  90.00  &  87.33  & \secondres{99.33} &  89.33  &  94.00  &  85.33  &  93.33  &  98.00  \\
Ham  & \boldres{80.00} &  51.43  &  65.71  &  69.43  & \secondres{78.10} &  73.33  &  77.14  &  75.00  &  72.38  \\
HandOutlines  & \secondres{88.90} &  36.20  &  86.30  & \boldres{94.35} &  86.00  &  85.20  &  86.40  &  85.81  &  85.70  \\
Haptics  & \secondres{47.73} &  26.95  &  37.01  & \boldres{50.84} &  43.83  &  41.23  &  46.10  &  44.06  &  43.51  \\
Herring  &  67.19  &  40.63  &  59.38  &  64.38  & \secondres{68.75} &  64.06  & \boldres{70.31} &  60.94  &  64.06  \\
InlineSkate  &  36.73  &  18.91  &  25.82  & \boldres{39.64} &  30.91  &  29.45  &  27.82  &  29.76  & \secondres{38.55} \\
InsectWingbeatSound  & \secondres{66.36} &  63.23  &  60.00  &  63.92  &  64.29  &  57.83  &  64.75  & \boldres{66.52} &  63.79  \\
ItalyPowerDemand  &  97.08  &  96.89  &  97.08  & \secondres{97.17} & \boldres{97.28} &  96.60  &  96.89  &  96.44  &  95.63  \\
LargeKitchenAppliances  & \secondres{81.87} &  33.33  &  47.20  &  81.47  &  53.87  &  63.20  &  42.13  &  76.08  & \boldres{86.40} \\
Lighting2  & \secondres{83.61} &  54.10  &  72.13  &  73.61  &  75.41  &  75.41  &  67.21  &  73.56  & \boldres{86.89} \\
Lighting7  & \boldres{83.56} &  53.42  &  72.60  &  68.63  &  72.60  &  67.12  &  64.38  &  81.53  & \secondres{83.56} \\
MALLAT  & \secondres{94.71} &  91.86  &  54.50  &  94.12  &  93.48  &  84.01  & \boldres{95.05} &  91.11  &  89.13  \\
Meat  & \boldres{93.33} &  50.00  &  33.33  & \secondres{93.33} &  88.33  &  91.67  &  80.00  &  31.25  &  91.67  \\
MedicalImages  &  72.76  &  61.18  &  58.95  & \secondres{75.42} &  65.79  &  63.03  &  59.61  &  74.35  & \boldres{75.79} \\
MiddlePhalanxOutlineAgeGroup  & \secondres{81.25} &  74.50  &  78.75  & \boldres{83.64} &  80.75  &  79.75  &  80.75  &  78.25  &  75.25  \\
MiddlePhalanxOutlineCorrect  & \boldres{84.00} &  64.67  &  64.67  &  61.36  &  64.50  &  64.83  &  64.50  &  52.47  & \secondres{71.67} \\
MiddlePhalanxTW  & \boldres{65.91} &  64.91  &  64.66  &  53.77  &  64.66  &  64.16  & \secondres{65.16} &  56.10  &  61.65  \\
MoteStrain  & \boldres{93.13} &  87.14  &  88.34  &  83.49  &  87.22  & \secondres{89.54} &  86.74  &  85.28  &  87.86  \\
NonInvasiveFatalECG\_Thorax1  & \secondres{93.44} &  72.98  &  81.58  & \boldres{95.65} &  86.97  &  78.73  &  92.98  &  84.58  &  90.48  \\
NonInvasiveFatalECG\_Thorax2  & \secondres{93.74} &  88.04  &  84.38  & \boldres{95.59} &  90.53  &  85.24  &  93.49  &  82.50  &  93.74  \\
OliveOil  &  40.00  &  40.00  &  40.00  &  80.33  &  60.00  & \secondres{83.33} &  70.00  &  42.86  & \boldres{90.00} \\
OSULeaf  &  74.79  &  9.50  &  43.39  & \boldres{82.89} &  49.59  &  42.15  &  45.04  &  63.28  & \secondres{76.86} \\
PhalangesOutlinesCorrect  & \secondres{82.40} &  77.04  &  68.30  & \boldres{83.11} &  69.35  &  65.97  &  66.90  &  78.73  &  80.77  \\
Phoneme  & \secondres{27.27} &  3.22  &  9.70  &  20.92  &  11.23  &  9.12  &  8.60  & \boldres{30.04} &  26.79  \\
Plane  & \boldres{100.00} &  97.14  &  98.10  & \secondres{100.00} &  98.10  &  99.05  &  97.14  &  96.43  &  100.00  \\
ProximalPhalanxOutlineAgeGroup  & \secondres{88.29} &  83.90  &  86.34  & \boldres{90.17} &  86.34  &  86.34  &  85.85  &  73.34  &  81.95  \\
ProximalPhalanxOutlineCorrect  & \boldres{91.75} &  81.79  &  77.66  &  86.59  &  84.54  &  78.01  &  81.79  &  87.17  & \secondres{87.29} \\
ProximalPhalanxTW  & \boldres{83.00} &  81.50  &  81.75  &  78.98  &  80.00  &  80.25  & \secondres{82.75} &  72.75  &  79.00  \\
RefrigerationDevices  & \boldres{55.47} &  33.60  &  33.60  &  50.40  &  42.40  &  45.87  &  38.67  &  49.74  & \secondres{51.20} \\
ScreenType  & \secondres{44.80} &  37.07  &  44.00  &  41.55  & \boldres{45.07} &  44.80  &  40.27  &  39.99  &  40.00  \\
ShapeletSim  & \boldres{90.00} &  49.44  &  50.00  &  65.72  &  57.22  &  56.67  &  56.67  &  61.98  & \secondres{87.78} \\
ShapesAll  &  85.17  &  61.17  &  64.33  & \secondres{86.63} &  68.17  &  61.00  &  61.83  &  79.11  & \boldres{88.00} \\
SmallKitchenAppliances  & \boldres{76.27} &  33.33  &  45.60  &  62.13  &  55.47  &  61.60  &  41.33  & \secondres{74.74} &  71.20  \\
SonyAIBORobotSurface  &  85.86  &  42.93  &  70.55  & \boldres{93.16} &  81.03  &  83.69  &  70.38  &  68.46  & \secondres{89.18} \\
SonyAIBORobotSurfaceII  & \boldres{92.44} &  70.30  &  85.62  & \secondres{91.26} &  85.73  &  85.73  &  85.10  &  86.15  &  90.66  \\
StarLightCurves  & \secondres{97.41} &  92.70  &  89.22  & \boldres{97.63} &  92.74  &  86.09  &  91.96  &  96.80  &  96.28  \\
Strawberry  & \boldres{98.37} &  94.94  &  93.15  & \secondres{97.84} &  94.45  &  93.15  &  95.76  &  93.59  &  96.57  \\
SwedishLeaf  & \boldres{96.16} &  88.32  &  83.40  & \secondres{96.10} &  82.08  &  76.64  &  80.96  &  92.31  &  93.60  \\
Symbols  &  94.07  &  16.98  &  86.43  & \boldres{96.71} &  86.13  &  82.21  &  84.72  &  86.08  & \secondres{96.58} \\
Synthetic\_control  & \boldres{100.00} &  97.67  & \secondres{99.67} &  99.53  &  93.67  &  99.67  &  87.00  &  99.67  &  99.67  \\
ToeSegmentation1  &  87.72  &  52.63  &  61.40  & \boldres{94.21} &  62.28  &  66.23  &  60.09  &  78.75  & \secondres{92.11} \\
ToeSegmentation2  & \secondres{90.00} &  75.38  &  86.15  & \boldres{91.00} &  81.54  &  76.92  &  58.46  &  59.72  &  87.69  \\
Trace  & \boldres{100.00} &  68.00  &  66.00  & \secondres{100.00} &  74.00  &  100.00  &  67.00  &  97.32  &  100.00  \\
TwoLeadECG  &  93.85  &  76.56  &  68.74  & \boldres{100.00} &  86.65  &  84.55  &  91.48  &  81.63  & \secondres{99.78} \\
Two\_Patterns  & \boldres{100.00} &  99.58  &  98.00  & \secondres{100.00} &  79.90  &  93.23  &  84.13  &  100.00  &  99.21  \\
uWaveGestureLibrary\_X  & \boldres{82.80} &  69.65  &  69.37  & \secondres{82.64} &  66.78  &  65.02  &  64.77  &  80.97  &  77.89  \\
uWaveGestureLibrary\_Y  & \secondres{73.53} &  54.22  &  62.67  & \boldres{73.83} &  61.33  &  55.11  &  60.19  &  71.22  &  67.87  \\
uWaveGestureLibrary\_Z  & \boldres{75.15} &  59.27  &  60.44  & \secondres{75.05} &  59.35  &  55.05  &  56.98  &  72.92  &  72.67  \\
uWaveGestureLibraryAll  & \boldres{97.57} &  85.54  &  90.28  & \secondres{97.20} &  88.02  &  87.58  &  88.22  &  96.54  &  91.99  \\
wafer  &  99.81  &  99.58  &  99.71  & \secondres{99.84} &  98.39  &  99.63  &  94.78  &  99.69  & \boldres{99.85} \\
Wine  &  66.67  &  53.70  &  50.00  &  71.30  &  68.52  & \secondres{77.78} &  72.22  &  57.81  & \boldres{85.19} \\
WordsSynonyms  & \secondres{69.28} &  7.37  &  50.16  & \boldres{71.30} &  56.90  &  49.06  &  44.83  &  66.12  &  68.81  \\
Worms  & \secondres{60.77} &  17.68  &  43.65  & \boldres{65.97} &  34.25  &  34.81  &  31.49  &  51.98  &  55.80  \\
WormsTwoClass  & \boldres{77.35} &  58.01  &  62.98  & \secondres{76.62} &  62.43  &  60.77  &  58.01  &  64.48  &  69.61  \\
yoga  & \secondres{85.83} &  71.83  &  67.77  & \boldres{90.49} &  73.87  &  68.43  &  65.13  &  77.46  &  84.23  \\ \midrule
Average  & \boldres{83.18} &  61.58  &  65.27  & \secondres{81.42} &  73.47  &  71.84  &  69.68  &  75.07  &  81.42  \\
        \midrule
        1st count & \boldres{38} & 0 & 1 & \secondres{27} & 2 & 2 & 2 & 4 & 9 \\ 
        \bottomrule
    \caption{Full classification results on the UCR datasets in terms of accuracy (as \%).}
    \label{tbl:clf_ucr_full}
\end{longtable}
}

\begin{table}[!ht]
    \centering
    \caption{Full classification results on the UEA datasets in terms of accuracy (as \%).}
    \label{tbl:clf_uea_full}
    \scalebox{0.85}{
    \begin{NiceTabular}{l|c|c|c|c|c|c|c|c|c}
    \toprule
        Dataset & \abb & GPT4TS & TimesNet & ROCKET & CrossF. & PatchTST & MLP & TS-TCC & TS2VEC\\  \midrule

ArticularyWordRecognition  & \secondres{99.00} &  93.33  &  96.18  & \boldres{99.33} &  98.00  &  97.67  &  97.33  &  98.00  &  87.33  \\
AtrialFibrillation  &  40.00  &  33.33  &  33.33  &  20.00  &  46.66  & \boldres{53.33} &  46.66  &  33.33  & \secondres{53.33} \\
BasicMotions  & \boldres{100.00} &  92.50  & \secondres{100.00} &  100.00  &  90.00  &  92.50  &  85.00  &  100.00  &  92.50  \\
Cricket  & \boldres{98.61} &  8.33  &  87.50  & \secondres{98.61} &  84.72  &  84.72  &  91.67  &  93.06  &  65.28  \\
Epilepsy  & \boldres{98.55} &  85.51  &  78.13  & \secondres{98.55} &  73.19  &  65.94  &  60.14  &  97.10  &  62.32  \\
EthanolConcentration  &  30.42  &  25.48  &  27.73  & \boldres{42.58} &  34.98  &  28.90  &  33.46  &  32.32  & \secondres{40.68} \\
FaceDetection  &  66.77  &  65.58  & \secondres{67.47} &  64.70  &  66.17  & \boldres{68.96} &  67.42  &  63.05  &  50.96  \\
FingerMovements  &  61.00  &  57.00  &  59.38  &  61.00  & \boldres{64.00} &  62.00  & \secondres{64.00} &  44.00  &  51.00  \\
HandMovementDirection  &  52.70  &  18.92  &  50.00  &  50.00  & \secondres{58.11} &  58.11  &  58.11  & \boldres{64.86} &  32.43  \\
Handwriting  & \boldres{57.88} &  3.76  &  26.18  & \secondres{48.47} &  26.24  &  26.00  &  22.47  &  47.76  &  15.53  \\
Heartbeat  & \boldres{77.56} &  36.59  &  74.48  &  69.76  &  76.59  &  76.59  &  73.17  & \secondres{77.07} &  69.76  \\
InsectWingbeat  & \boldres{10.00} & \secondres{10.00} &  10.00  &  10.00  &  10.00  &  10.00  &  10.00  &  10.00  &  10.00  \\
JapaneseVowels  & \boldres{99.19} &  98.11  &  97.83  &  95.68  & \secondres{98.92} &  98.65  &  97.84  &  97.30  &  90.00  \\
Libras  & \boldres{92.78} &  79.44  &  77.84  &  83.89  &  76.11  &  81.11  &  73.33  & \secondres{86.67} &  85.56  \\
LSST  & \secondres{66.34} &  46.39  &  59.21  &  54.10  &  42.82  & \boldres{67.80} &  35.77  &  49.23  &  39.01  \\
MotorImagery  & \boldres{62.00} &  50.00  &  51.04  &  53.00  & \secondres{61.00} &  61.00  &  61.00  &  47.00  &  47.00  \\
NATOPS  &  95.56  &  91.67  &  81.82  &  83.33  &  88.33  & \boldres{96.67} &  93.89  & \secondres{96.11} &  82.22  \\
PEMS-SF  &  83.82  &  87.28  & \secondres{88.13} &  75.10  &  82.08  & \boldres{88.44} &  82.08  &  86.71  &  72.25  \\
PenDigits  & \secondres{98.94} &  97.74  &  98.19  &  97.34  &  93.65  & \boldres{99.23} &  92.94  &  98.51  &  97.40  \\
PhonemeSpectra  &  17.75  &  3.01  & \secondres{18.24} &  17.60  &  7.55  &  11.69  &  7.10  & \boldres{25.92} &  8.23  \\
RacketSports  & \boldres{90.79} &  76.97  &  82.64  & \secondres{86.18} &  81.58  &  84.21  &  78.95  &  84.87  &  74.34  \\
SelfRegulationSCP1  & \secondres{91.81} &  91.47  &  77.43  &  84.64  & \boldres{92.49} &  89.76  &  88.40  &  91.13  &  77.13  \\
SelfRegulationSCP2  & \boldres{61.67} &  51.67  &  52.84  & \secondres{54.44} &  53.33  &  54.44  &  51.67  &  53.89  &  51.11  \\
SpokenArabicDigits  & \boldres{99.91} &  99.36  &  98.36  &  99.20  &  96.41  &  99.68  &  96.68  & \secondres{99.77} &  85.27  \\
StandWalkJump  &  46.67  &  33.33  &  53.33  &  46.67  &  53.33  & \boldres{60.00} & \secondres{60.00} &  40.00  &  46.67  \\
UWaveGestureLibrary  & \secondres{91.25} &  84.38  &  83.13  & \boldres{94.40} &  81.56  &  80.00  &  81.88  &  86.25  &  62.81  \\ \midrule
Average  & \boldres{72.73} &  58.51  &  66.55  &  68.79  &  66.84  &  69.13  &  65.81  & \secondres{69.38} &  59.62  \\
        \midrule
        1st count & \boldres{12} & 0 & 0 & 3 & 2 & \secondres{7} & 0 & 2 & 0 \\ 
        \bottomrule
    \end{NiceTabular}
    }
\end{table}

\begin{table}[!ht]
    \centering
    \caption{Full classification results on the human activity recognition and biomedical signal datasets in terms of accuracy (as \%).}
    \label{tbl:clf_others_full}
    \begin{tabular}{l|c|c|c|c|c|c|c|c|c}
    \toprule
        Dataset & \abb & GPT4TS & TimesNet & ROCKET & CrossF. & PatchTST & MLP & TS-TCC & TS2VEC \\ \midrule
        UCIHAR  & \secondres{96.06} &  91.24  &  91.34  &  94.37  &  76.59  &  92.70  &  63.49  &  95.95  & \boldres{96.19} \\
WISDM  & \boldres{97.77} &  89.49  &  89.61  &  97.03  &  77.31  &  95.94  &  58.88  & \secondres{97.05} &  93.87  \\
HHAR  & \boldres{98.53} &  97.40  &  93.59  &  97.93  &  78.74  &  95.96  &  47.70  & \secondres{98.49} &  97.05  \\ \midrule
Average  & \boldres{97.46} &  92.71  &  91.51  &  96.44  &  77.55  &  94.87  &  56.69  & \secondres{97.16} &  95.70  \\
        \bottomrule
        \toprule
        EEG  & \secondres{82.10} &  76.37  &  75.86  &  76.69  &  53.30  &  69.69  &  49.70  & \boldres{86.06} &  75.13  \\
ECG  & \secondres{98.37} &  97.70  &  98.33  &  97.72  &  88.33  &  98.06  &  91.57  & \boldres{98.44} &  97.48  \\ \midrule
Average  & \secondres{90.24} &  87.04  &  87.10  &  87.20  &  70.82  &  83.87  &  70.63  & \boldres{92.25} &  86.31  \\ \bottomrule
    \end{tabular}
\end{table}

\subsection{Forecasting}
\begin{table*}[!h]
\caption{Full forecasting results on different prediction lengths $\in \{96, 192, 336, 720\}$. Lower MSE indicates better performance.}
\label{tbl:forecasting_full}
\begin{center}
\begin{small}
\scalebox{0.70}{
\setlength\tabcolsep{3pt}
\begin{tabular}{c|c|cc|cc|cc|cc|cc|cc|cc|cc|cc|cc|cc|cc}
\toprule

\multicolumn{2}{c}{Methods}&\multicolumn{2}{c}{\abb}&\multicolumn{2}{c}{Time-LLM}&\multicolumn{2}{c}{iTransformer}&\multicolumn{2}{c}{PatchTST}&\multicolumn{2}{c}{Crossformer}&\multicolumn{2}{c}{FEDformer}&\multicolumn{2}{c}{Autoformer}&\multicolumn{2}{c}{RLinear}&\multicolumn{2}{c}{Dlinear}&\multicolumn{2}{c}{TimesNet}&\multicolumn{2}{c}{GPT4TS}&\multicolumn{2}{c}{SCINet} \\

\midrule

\multicolumn{2}{c|}{Metric} & MSE  & MAE & MSE  & MAE & MSE & MAE& MSE & MAE& MSE  & MAE& MSE  & MAE& MSE  & MAE& MSE  & MAE& MSE  & MAE& MSE  & MAE& MSE  & MAE& MSE  & MAE \\
\midrule

\multirow{5}{*}{\rotatebox{90}{$ECL$}}
~  &  96  & \secondres{0.136} & \secondres{0.229} & \boldres{0.131} & \boldres{0.224} &  0.148  &  0.240  &  0.138  &  0.230  &  0.219  &  0.314  &  0.193  &  0.308  &  0.201  &  0.317  &  0.201  &  0.281  &  0.140  &  0.237  &  0.168  &  0.272  &  0.139  &  0.238  &  0.247  &  0.345  \\
~  &  192  & \secondres{0.152} &  0.244  &  0.152  & \boldres{0.241} &  0.162  &  0.253  & \boldres{0.149} & \secondres{0.243} &  0.231  &  0.322  &  0.201  &  0.315  &  0.222  &  0.334  &  0.201  &  0.283  &  0.153  &  0.249  &  0.184  &  0.289  &  0.153  &  0.251  &  0.257  &  0.355  \\
~  &  336  & \secondres{0.168} & \secondres{0.262} & \boldres{0.160} & \boldres{0.248} &  0.178  &  0.269  &  0.169  &  0.262  &  0.246  &  0.337  &  0.214  &  0.329  &  0.231  &  0.338  &  0.215  &  0.298  &  0.169  &  0.267  &  0.198  &  0.300  &  0.169  &  0.266  &  0.269  &  0.369  \\
~  &  720  &  0.205  & \boldres{0.293} & \boldres{0.192} &  0.298  &  0.225  &  0.317  &  0.211  &  0.299  &  0.280  &  0.363  &  0.246  &  0.355  &  0.254  &  0.361  &  0.257  &  0.331  & \secondres{0.203} &  0.301  &  0.220  &  0.320  &  0.206  & \secondres{0.297} &  0.299  &  0.390  \\ \cmidrule{2-25}
~  &  Avg  & \secondres{0.165} & \secondres{0.257} & \boldres{0.158} & \boldres{0.252} &  0.178  &  0.270  &  0.167  &  0.259  &  0.244  &  0.334  &  0.214  &  0.327  &  0.227  &  0.338  &  0.219  &  0.298  &  0.166  &  0.264  &  0.193  &  0.295  &  0.167  &  0.263  &  0.268  &  0.365  \\

\midrule

\multirow{5}{*}{\rotatebox{90}{$ETTh1$}}
~  &  96  & \secondres{0.370} & \secondres{0.394} & \boldres{0.362} & \boldres{0.392} &  0.386  &  0.405  &  0.382  &  0.401  &  0.423  &  0.448  &  0.376  &  0.419  &  0.449  &  0.459  &  0.386  &  0.395  &  0.375  &  0.399  &  0.384  &  0.402  &  0.376  &  0.397  &  0.654  &  0.599  \\
~  &  192  &  0.412  & \secondres{0.417} & \boldres{0.398} &  0.418  &  0.441  &  0.436  &  0.428  &  0.425  &  0.471  &  0.474  &  0.420  &  0.448  &  0.500  &  0.482  &  0.437  &  0.424  & \secondres{0.405} & \boldres{0.416} &  0.436  &  0.429  &  0.416  &  0.418  &  0.719  &  0.631  \\
~  &  336  & \boldres{0.399} & \boldres{0.416} & \secondres{0.430} & \secondres{0.427} &  0.487  &  0.458  &  0.451  &  0.436  &  0.570  &  0.546  &  0.459  &  0.465  &  0.521  &  0.496  &  0.479  &  0.446  &  0.439  &  0.443  &  0.491  &  0.469  &  0.442  &  0.433  &  0.778  &  0.659  \\
~  &  720  &  0.472  &  0.475  & \boldres{0.442} & \secondres{0.457} &  0.503  &  0.491  & \secondres{0.452} &  0.459  &  0.653  &  0.621  &  0.506  &  0.507  &  0.514  &  0.512  &  0.481  &  0.470  &  0.472  &  0.490  &  0.521  &  0.500  &  0.477  & \boldres{0.456} &  0.836  &  0.699  \\ \cmidrule{2-25}
~  &  Avg  & \secondres{0.413} & \secondres{0.426} & \boldres{0.408} & \boldres{0.423} &  0.454  &  0.448  &  0.428  &  0.430  &  0.529  &  0.522  &  0.440  &  0.460  &  0.496  &  0.487  &  0.446  &  0.434  &  0.423  &  0.437  &  0.458  &  0.450  &  0.428  &  0.426  &  0.747  &  0.647  \\

\midrule

\multirow{5}{*}{\rotatebox{90}{$ETTh2$}}
~  &  96  & \secondres{0.280} &  0.341  & \boldres{0.268} & \boldres{0.328} &  0.297  &  0.349  &  0.285  &  0.340  &  0.745  &  0.584  &  0.358  &  0.397  &  0.346  &  0.388  &  0.288  & \secondres{0.338} &  0.289  &  0.353  &  0.340  &  0.374  &  0.285  &  0.342  &  0.707  &  0.621  \\
~  &  192  & \secondres{0.330} & \boldres{0.375} & \boldres{0.329} & \secondres{0.375} &  0.380  &  0.400  &  0.356  &  0.386  &  0.877  &  0.656  &  0.429  &  0.439  &  0.456  &  0.452  &  0.374  &  0.390  &  0.383  &  0.418  &  0.402  &  0.414  &  0.354  &  0.389  &  0.860  &  0.689  \\
~  &  336  & \boldres{0.317} & \boldres{0.374} &  0.368  &  0.409  &  0.428  &  0.432  & \secondres{0.350} & \secondres{0.395} &  1.043  &  0.731  &  0.496  &  0.487  &  0.482  &  0.486  &  0.415  &  0.426  &  0.448  &  0.465  &  0.452  &  0.452  &  0.373  &  0.407  &  1.000  &  0.744  \\
~  &  720  &  0.404  &  0.440  & \boldres{0.372} & \boldres{0.420} &  0.427  &  0.445  & \secondres{0.395} & \secondres{0.427} &  1.104  &  0.763  &  0.463  &  0.474  &  0.515  &  0.511  &  0.420  &  0.440  &  0.605  &  0.551  &  0.462  &  0.468  &  0.406  &  0.441  &  1.249  &  0.838  \\ \cmidrule{2-25}
~  &  Avg  & \boldres{0.333} & \boldres{0.383} & \secondres{0.334} & \secondres{0.383} &  0.383  &  0.407  &  0.347  &  0.387  &  0.942  &  0.684  &  0.437  &  0.449  &  0.450  &  0.459  &  0.374  &  0.399  &  0.431  &  0.447  &  0.414  &  0.427  &  0.355  &  0.395  &  0.954  &  0.723  \\

\midrule

\multirow{5}{*}{\rotatebox{90}{$ETTm1$}}
~  &  96  & \secondres{0.289} &  0.349  & \boldres{0.272} & \boldres{0.334} &  0.334  &  0.368  &  0.291  & \secondres{0.340} &  0.404  &  0.426  &  0.379  &  0.419  &  0.505  &  0.475  &  0.355  &  0.376  &  0.299  &  0.343  &  0.338  &  0.375  &  0.292  &  0.346  &  0.418  &  0.438  \\
~  &  192  & \secondres{0.328} &  0.370  & \boldres{0.310} & \boldres{0.358} &  0.377  &  0.391  &  0.328  & \secondres{0.365} &  0.450  &  0.451  &  0.426  &  0.441  &  0.553  &  0.496  &  0.391  &  0.392  &  0.335  &  0.365  &  0.374  &  0.387  &  0.332  &  0.372  &  0.439  &  0.450  \\
~  &  336  & \secondres{0.355} &  0.389  & \boldres{0.352} & \boldres{0.384} &  0.426  &  0.420  &  0.365  &  0.389  &  0.532  &  0.515  &  0.445  &  0.459  &  0.621  &  0.537  &  0.424  &  0.415  &  0.369  & \secondres{0.386} &  0.410  &  0.411  &  0.366  &  0.394  &  0.490  &  0.485  \\
~  &  720  &  0.421  &  0.425  & \boldres{0.383} & \boldres{0.411} &  0.491  &  0.459  &  0.422  &  0.423  &  0.666  &  0.589  &  0.543  &  0.490  &  0.671  &  0.561  &  0.487  &  0.450  &  0.425  & \secondres{0.421} &  0.478  &  0.450  & \secondres{0.417} &  0.421  &  0.595  &  0.550  \\ \cmidrule{2-25}
~  &  Avg  & \secondres{0.348} &  0.383  & \boldres{0.329} & \boldres{0.372} &  0.407  &  0.410  &  0.352  & \secondres{0.379} &  0.513  &  0.495  &  0.448  &  0.452  &  0.588  &  0.517  &  0.414  &  0.408  &  0.357  &  0.379  &  0.400  &  0.406  &  0.352  &  0.383  &  0.486  &  0.481  \\

\midrule

\multirow{5}{*}{\rotatebox{90}{$ETTm2$}}
~  &  96  &  0.169  &  0.259  & \boldres{0.161} & \boldres{0.253} &  0.180  &  0.264  &  0.169  & \secondres{0.254} &  0.287  &  0.366  &  0.203  &  0.287  &  0.255  &  0.339  &  0.182  &  0.265  & \secondres{0.167} &  0.260  &  0.187  &  0.267  &  0.173  &  0.262  &  0.286  &  0.377  \\
~  &  192  & \secondres{0.224} &  0.297  & \boldres{0.219} & \boldres{0.293} &  0.250  &  0.309  &  0.230  & \secondres{0.294} &  0.414  &  0.492  &  0.269  &  0.328  &  0.281  &  0.340  &  0.246  &  0.304  &  0.224  &  0.303  &  0.249  &  0.309  &  0.229  &  0.301  &  0.399  &  0.445  \\
~  &  336  & \secondres{0.275} & \boldres{0.329} & \boldres{0.271} & \secondres{0.329} &  0.311  &  0.348  &  0.280  &  0.329  &  0.597  &  0.542  &  0.325  &  0.366  &  0.339  &  0.372  &  0.307  &  0.342  &  0.281  &  0.342  &  0.321  &  0.351  &  0.286  &  0.341  &  0.637  &  0.591  \\
~  &  720  & \secondres{0.354} & \secondres{0.380} & \boldres{0.352} & \boldres{0.379} &  0.412  &  0.407  &  0.378  &  0.386  &  1.730  &  1.042  &  0.421  &  0.415  &  0.433  &  0.432  &  0.407  &  0.398  &  0.397  &  0.421  &  0.408  &  0.403  &  0.378  &  0.401  &  0.960  &  0.735  \\ \cmidrule{2-25}
~  &  Avg  & \secondres{0.256} & \secondres{0.316} & \boldres{0.251} & \boldres{0.313} &  0.288  &  0.332  &  0.264  &  0.316  &  0.757  &  0.611  &  0.305  &  0.349  &  0.327  &  0.371  &  0.286  &  0.327  &  0.267  &  0.332  &  0.291  &  0.333  &  0.267  &  0.326  &  0.571  &  0.537  \\

\midrule

\multirow{5}{*}{\rotatebox{90}{$Exchange$}}
~  &  96  &  0.083  & \secondres{0.201} & - & - &  0.086  &  0.206  &  0.088  &  0.205  &  0.256  &  0.367  &  0.148  &  0.278  &  0.197  &  0.323  &  0.093  &  0.217  & \boldres{0.081} &  0.203  &  0.107  &  0.234  & \secondres{0.082} & \boldres{0.199} &  0.267  &  0.396  \\
~  &  192  &  0.177  &  0.299  & - & - &  0.177  &  0.299  &  0.176  &  0.299  &  0.470  &  0.509  &  0.271  &  0.315  &  0.300  &  0.369  &  0.184  &  0.307  & \boldres{0.157} & \boldres{0.293} &  0.226  &  0.344  & \secondres{0.171} & \secondres{0.293} &  0.351  &  0.459  \\
~  &  336  &  0.331  &  0.417  & - & - &  0.331  &  0.417  & \boldres{0.301} & \boldres{0.397} &  1.268  &  0.883  &  0.460  &  0.427  &  0.509  &  0.524  &  0.351  &  0.432  & \secondres{0.305} & \secondres{0.414} &  0.367  &  0.448  &  0.354  &  0.428  &  1.324  &  0.853  \\
~  &  720  &  0.888  &  0.739  & - & - & \secondres{0.847} & \secondres{0.691} &  0.901  &  0.714  &  1.767  &  1.068  &  1.195  &  0.695  &  1.447  &  0.941  &  0.886  &  0.714  & \boldres{0.643} & \boldres{0.601} &  0.964  &  0.746  &  0.877  &  0.704  &  1.058  &  0.797  \\ \cmidrule{2-25}
~  &  Avg  &  0.370  &  0.414  & - & - & \secondres{0.360} & \secondres{0.403} &  0.367  &  0.404  &  0.940  &  0.707  &  0.519  &  0.429  &  0.613  &  0.539  &  0.379  &  0.418  & \boldres{0.297} & \boldres{0.378} &  0.416  &  0.443  &  0.371  &  0.406  &  0.750  &  0.626  \\

\midrule

\multirow{5}{*}{\rotatebox{90}{$Traffic$}}
~  &  96  & \secondres{0.372} & \secondres{0.261} & \boldres{0.362} & \boldres{0.248} &  0.395  &  0.268  &  0.401  &  0.267  &  0.522  &  0.290  &  0.587  &  0.366  &  0.613  &  0.388  &  0.649  &  0.389  &  0.410  &  0.282  &  0.593  &  0.321  &  0.388  &  0.282  &  0.788  &  0.499  \\
~  &  192  & \secondres{0.388} & \secondres{0.266} & \boldres{0.374} & \boldres{0.247} &  0.417  &  0.276  &  0.406  &  0.268  &  0.530  &  0.293  &  0.604  &  0.373  &  0.616  &  0.382  &  0.601  &  0.366  &  0.423  &  0.287  &  0.617  &  0.336  &  0.407  &  0.290  &  0.789  &  0.505  \\
~  &  336  & \secondres{0.394} & \boldres{0.269} & \boldres{0.385} & \secondres{0.271} &  0.433  &  0.283  &  0.421  &  0.277  &  0.558  &  0.305  &  0.621  &  0.383  &  0.622  &  0.337  &  0.609  &  0.369  &  0.436  &  0.296  &  0.629  &  0.336  &  0.412  &  0.294  &  0.797  &  0.508  \\
~  &  720  & \boldres{0.430} & \secondres{0.289} & \secondres{0.43} & \boldres{0.288} &  0.467  &  0.302  &  0.452  &  0.297  &  0.589  &  0.328  &  0.626  &  0.382  &  0.660  &  0.408  &  0.647  &  0.387  &  0.466  &  0.315  &  0.640  &  0.350  &  0.450  &  0.312  &  0.841  &  0.523  \\ \cmidrule{2-25}
~  &  Avg  & \secondres{0.396} & \secondres{0.271} & \boldres{0.388} & \boldres{0.264} &  0.428  &  0.282  &  0.420  &  0.277  &  0.550  &  0.304  &  0.610  &  0.376  &  0.628  &  0.379  &  0.627  &  0.378  &  0.434  &  0.295  &  0.620  &  0.336  &  0.414  &  0.295  &  0.804  &  0.509  \\

\midrule

\multirow{5}{*}{\rotatebox{90}{$Weather$}}
~  &  96  & \secondres{0.148} & \boldres{0.197} & \boldres{0.147} & \secondres{0.201} &  0.174  &  0.214  &  0.160  &  0.204  &  0.158  &  0.230  &  0.217  &  0.296  &  0.266  &  0.336  &  0.192  &  0.232  &  0.176  &  0.237  &  0.172  &  0.220  &  0.162  &  0.212  &  0.221  &  0.306  \\
~  &  192  & \secondres{0.193} & \secondres{0.241} & \boldres{0.189} & \boldres{0.234} &  0.221  &  0.254  &  0.204  &  0.245  &  0.206  &  0.277  &  0.276  &  0.336  &  0.307  &  0.367  &  0.240  &  0.271  &  0.220  &  0.282  &  0.219  &  0.261  &  0.204  &  0.248  &  0.261  &  0.340  \\
~  &  336  & \boldres{0.245} & \secondres{0.282} &  0.262  & \boldres{0.279} &  0.278  &  0.296  &  0.257  &  0.285  &  0.272  &  0.335  &  0.339  &  0.380  &  0.359  &  0.395  &  0.292  &  0.307  &  0.265  &  0.319  &  0.280  &  0.306  & \secondres{0.254} &  0.286  &  0.309  &  0.378  \\
~  &  720  &  0.325  & \secondres{0.337} & \boldres{0.304} & \boldres{0.316} &  0.358  &  0.349  &  0.329  &  0.338  &  0.398  &  0.418  &  0.403  &  0.428  &  0.419  &  0.428  &  0.364  &  0.353  & \secondres{0.323} &  0.362  &  0.365  &  0.359  &  0.326  &  0.337  &  0.377  &  0.427  \\ \cmidrule{2-25}
~  &  Avg  & \secondres{0.228} & \secondres{0.264} & \boldres{0.225} & \boldres{0.257} &  0.258  &  0.278  &  0.238  &  0.268  &  0.259  &  0.315  &  0.309  &  0.360  &  0.338  &  0.382  &  0.272  &  0.291  &  0.246  &  0.300  &  0.259  &  0.287  &  0.237  &  0.271  &  0.292  &  0.363  \\

\bottomrule
\end{tabular}
}
\end{small}
\end{center}
\vskip -0.1in
\end{table*}

\subsection{Anomaly Detection}
\begin{table}[!h]
\caption{Full results for the anomaly detection.}
\label{tbl:anomaly_full}
\begin{center}
\begin{small}
\scalebox{0.79}{
\begin{threeparttable}[b]
\begin{tabular}{c|ccc|ccc|ccc|ccc|ccc|c}
\toprule

Methods &
\multicolumn{3}{c}{SMD} & \multicolumn{3}{c}{MSL} & \multicolumn{3}{c}{SMAP}& \multicolumn{3}{c}{SWaT} &\multicolumn{3}{c}{PSM} & Avg F1 \\
Metrics&P&R&F1&P&R&F1&P&R&F1&P&R&F1&P&R&F1&\%  \\

\midrule
\abb \textbf{\textit{(Ours)}}& 85.58&\boldres{90.37}&\boldres{87.91}& 77.46&\boldres{90.12}&83.32 & \secondres{92.45}&\boldres{64.47}&\boldres{75.96} & \secondres{91.50}&94.14&92.80 & 98.36 & \boldres{98.55}&\boldres{97.73} & \boldres{87.54} \\
GPT4TS&\secondres{88.89}&\secondres{84.98}&\secondres{86.89}&82.00&82.91&82.45&90.60&\secondres{60.95}&\secondres{72.88}&\boldres{92.20}&96.34&\boldres{94.23}&98.62&95.68&97.13&\secondres{86.72}\\
TimesNet &87.91&81.54&84.61&\boldres{89.54}&75.36&81.84&90.14&56.40&69.39&90.75&95.40&93.02&98.51&96.20&\secondres{97.34}&85.24\\
PatchTST&87.26&82.14&84.62&\secondres{88.34}&70.96&78.70&90.64&55.46&68.82&91.10&80.94&85.72&98.84&93.47&96.08&82.79\\
ETSformer&87.44&79.23&83.13&85.13&84.93&\boldres{85.03}&92.25&55.75&69.50&90.02&80.36&84.91&\boldres{99.31}&85.28&91.76&82.87\\
FEDformer&87.95&82.39&85.08&77.14&80.07&78.57&90.47&58.10&70.76&90.17&96.42&93.19&97.31&97.16&97.23&84.97\\
LightTS&87.10&78.42&82.53&82.40&75.78&78.95&\boldres{92.58}&55.27&69.21&91.98&94.72&\secondres{93.33}&98.37&95.97&97.15&84.23 \\
DLinear&83.62&71.52&77.10&84.34&85.42&\secondres{84.88}&92.32&55.41&69.26&80.91&95.30&87.52&98.28&89.26&93.55&82.46 \\
Stationary&88.33&81.21&84.62&68.55&\secondres{89.14}&77.50&89.37&59.02&71.09&68.03&96.75&79.88&97.82&96.76&97.29&82.08 \\
Autoformer&88.06&82.35&85.11&77.27&80.92&79.05&90.40&58.62&71.12&89.85&95.81&92.74&\secondres{99.08}&88.15&93.29&84.26 \\
Pyraformer&85.61&80.61&83.04&83.81&85.93&84.86&92.34&57.71&71.09&87.92&96.00&91.78&71.67&96.02&82.08&82.57 \\
Anomaly Transformer&\boldres{88.91}&82.23&85.49&79.61&87.37&83.31&91.85&58.11&71.18&72.51&\boldres{97.32}&83.10&68.35&94.72&79.40&80.50 \\
Informer&86.60&77.23&81.65&81.77&86.48&84.06&90.11&57.13&69.92&70.29&96.75&81.43&64.27&96.33&77.10&78.83 \\
Reformer&82.58&69.24&75.32&85.51&83.31&84.40&90.91&57.44&70.40&72.50&96.53&82.80&59.93&95.38&73.61&77.31 \\
LogTransformer&83.46&70.13&76.21&73.05&87.37&79.57&89.15&57.59&69.97&68.67&\secondres{97.31}&80.52&63.06&\secondres{98.00}&76.74&76.60 \\
Transformer&83.58&76.13&79.56&71.57&87.37&78.68&89.37&57.12&69.70&68.84&96.53&80.37&62.75&96.56&76.07&76.88 \\

\bottomrule

\end{tabular}

\end{threeparttable}

}
\end{small}
\end{center}
\vskip -0.1in
\end{table}

\section{Future Work}
\abb is aimed to be a foundation model for time series analysis. Therefore, we have some future directions toward achieving this goal. These are summarized as follows.

\paragraph{Large-Scale Pretraining}
We aim to explore the potential of TSLANet when pretrained on a diverse and large cohort of datasets. This would enable us to assess the model's generalization capabilities and its performance on few-shot and zero-shot learning tasks. In addition, this would give our model an advantage in competing against LLM-pretrained models in time series analysis.

\paragraph{Better Pretraining Task}
We aim to develop other pretraining tasks beyond the current masking approach, which, while straightforward and effective for initial learning, presents limitations in fully capturing the complexity of time series data. Masking may not adequately challenge the model to learn the intricate temporal dependencies and patterns essential for advanced classification and forecasting. This exploration will contribute to evolving TSLANet into a more refined and capable foundation model for time series analysis.

\paragraph{Enhanced Noise Reduction Techniques}
Building upon the adaptive spectral filtering capabilities of TSLANet, future work could explore more sophisticated noise reduction techniques that can adapt to a wider variety of noise patterns and distributions, as well as be adept to the quick fluctuations in short-term forecasting problems.
\end{document}